\documentclass{article}

\usepackage{microtype}
\usepackage{graphicx}
\usepackage{subcaption}
\usepackage{booktabs} 

\usepackage{hyperref}

\usepackage[accepted]{icml2023}

\usepackage{amsmath}
\usepackage{amssymb}
\usepackage{mathtools}
\usepackage{amsthm}

\usepackage{multirow}
\usepackage{physics}
\usepackage{enumitem}
\usepackage{bm}
\usepackage{xcolor}
\newcommand{\matr}[1]{\mathbf{#1}}

\definecolor{mycyan}{rgb}{0., 1,, 1.}
\definecolor{mymagenta}{rgb}{1., 0., 1.}

\usepackage[capitalize,noabbrev]{cleveref}

\theoremstyle{plain}

\theoremstyle{definition}

\theoremstyle{remark}

\usepackage[textsize=tiny]{todonotes}

\icmltitlerunning{Anomaly Detection in Networks via Score-Based Generative Models}

\begin{document}

\twocolumn[
\icmltitle{Anomaly Detection in Networks via Score-Based Generative Models}



\icmlsetsymbol{equal}{*}

\begin{icmlauthorlist}
\icmlauthor{Dmitrii Gavrilev}{skoltech}
\icmlauthor{Evgeny Burnaev}{skoltech,airi}
\end{icmlauthorlist}

\icmlaffiliation{skoltech}{Skolkovo Institute of Science and Technology, Moscow, Russia}
\icmlaffiliation{airi}{Artificial Intelligence Research Institute, Moscow, Russia}

\icmlcorrespondingauthor{Dmitrii Gavrilev}{dmitrygavrilyev@gmail.com}

\icmlkeywords{Machine Learning, ICML, Score-Based Modeling, Diffusion Models, Graph Neural Networks, Anomaly Detection}

\vskip 0.3in
]



\printAffiliationsAndNotice{}  

\begin{abstract}
Node outlier detection in attributed graphs is a challenging problem for which there is no method that would work well across different datasets. Motivated by the state-of-the-art results of score-based models in graph generative modeling, we propose to incorporate them into the aforementioned problem. Our method achieves competitive results on small-scale graphs. We provide an empirical analysis of the Dirichlet energy, and show that generative models might struggle to accurately reconstruct it.
\end{abstract}

\section{Introduction}

Graphs are a natural structure to represent various kinds of data, such as social networks, molecules, the Internet, and infrastructure networks, to name but a few. They describe the connectivity (relations) between objects. Analyzing networks might be crucial for both research and industry, e.g., for fraud detection. For instance, anomalous nodes might reveal fraudsters in a network of transactions, potentially preventing a significant loss of money \cite{gad-survey} or detecting false product reviews that could mislead customers \cite{rev2}. Another important scenario is recognizing the Out-Of-Distribution (OOD) nodes in a graph, on which a discriminative model may yield unreliable predictions \cite{energy-ood}.

A graph $G$ can be represented by the sets of nodes (vertices) $V$ and edges $E$. One of the ways to describe a graph is to build the adjacency matrix $\matr{A} \in \mathbb{R}^{n \times n}$, where $n$ is the number of nodes. Real-world networks are often attributed graphs, meaning that the nodes and edges might additionally have attributes described as vectors. In this paper, we use the notions of 'attributes' and 'features' interchangeably. Anomalous Node Detection (ANOS ND) is the problem of recognizing the nodes in a graph that deviate dramatically from the others \cite{gad-survey}. That is, the goal is to rank the nodes by the degree of abnormality. Due to the cost of labeling training datasets, unsupervised methods are preferred. However, to our knowledge, most of them fail to solve ANOS ND \cite{bond}. Hence, devising an unsupervised method that would work well across a wide variety of graphs remains challenging.

Recently, score-based generative modeling (or diffusion-based modeling) has been getting close attention due to its expressive generation. It has been incorporated in various modalities, including images \cite{diffbeatgans}, audio \cite{wavegrad}, video \cite{videodiffusion}, text \cite{diffuser}, and graphs \cite{scorebased, sde-graph, digress}. In our work, we leverage score-based graph generative models to detect anomalous nodes in a given attributed network. The behavior of a node can be characterized by its neighborhood, i.e., ego-graph. Formally, an ego-graph of a node $v$ is the induced subgraph of $v$ and its $k$-hop neighbors \cite{ego}. Our key idea is to view a network as a \emph{collection of ego-graphs}. This view allows us to learn the probability distribution induced by a network with score-based generative models. In the context of anomaly detection on networks, our contribution is twofold:
\begin{itemize}[nosep]
    \item Learning the distribution of ego-graphs with score-based modeling;
    \item Introducing measures of abnormality based on the reconstruction error of \emph{attributed} ego-graphs.
\end{itemize}

This paper is organized as follows. In Appendix \ref{appendix:review}, we review unsupervised methods for ANOS ND and the recent advancements in graph generation. In Section \ref{sec:gdss}, we briefly describe the training procedure of GDSS \emph{Graph Diffusion via the System of Stochastic Differential Equations} (GDSS \cite{sde-graph}). Next, in Section \ref{sec:methods}, we present our method of assigning anomaly scores. Section \ref{sec:experiments} describes the experimental setup, with Appendices \ref{appendix:datasets}-\ref{appendix:graphs} supplementing it. Section \ref{sec:discussion} concludes with a discussion of the results and provides insights regarding the limitations of our methods and directions for improvement.

\section{Background: Training GDSS}
\label{sec:gdss}
Let $\matr{G}_0=(\matr{X}_0, \matr{A}_0) \in \mathbb{R}^{N \times F} \times \mathbb{R}^{N \times N}$ be an attributed graph with node feature matrix $\matr{X}_0$ and adjacency matrix $\matr{A}_0$, where $N$ and $F$ denote the number of nodes and features, respectively. A graph $\matr{G}_0$ is an arbitrary ego-graph drawn from the distribution $p_{\textrm{data}}$ we want to learn. Forward diffusion of GDSS is a continuous process that destroys the graph structure and its properties. It is defined by the following Itô SDE:
\begin{equation}
\label{eq:forward}
    \dd{\matr{G}_t} = \matr{f}_t (\matr{G_t}) \dd{t} + g_t \dd{\matr{w}_t}, 
\end{equation}
where $\matr{f}_t$ and $g_t$ are the drift and diffusion functions, respectively, and $\matr{w}_t$ is the Wiener process. Note that $\matr{G}_t$, $\matr{f}_t$, $g_t$ and $\matr{w}_t$ are all functions of time $t$, which is written as a subscript for brevity. This process spans over a time horizon $[0, T]$. Therefore, $\matr{G}_0 \sim p_{\textrm{data}}$ denotes the original graph, whereas $\matr{G}_t \sim p_{0t}(\matr{G}_t | \matr{G}_0)$ indicates its terminal noisy version. Note that $\matr{G}_t$ at $t > 0$ is not a sparse graph since forward diffusion destroys the sparsity of the adjacency matrix.

Let the drift function be separable into linear attribute and adjacency components:
\begin{equation}
\label{eq:separability}
    \matr{f}_t(\matr{G}_t) = \left( \matr{f}_{1,t}(\matr{X}_t), \matr{f}_{2,t}(\matr{A}_t) \right).
\end{equation}
The choice of the linear drift terms allows us to factorize the transition kernel $p_{0t}(\matr{G}_t | \matr{G}_0)$:
\begin{equation}
    p_{0t}(\matr{G}_t | \matr{G}_0) = p_{0t}(\matr{X}_t | \matr{X}_0) p_{0t}(\matr{A}_t | \matr{A}_0).
\end{equation}
Moreover, sampling from $p_{0t}(\matr{X}_t | \matr{X}_0)$ and $p_{0t}(\matr{A}_t | \matr{A}_0)$ is fast because they are Gaussian, each with a known mean and covariance (see Sections 5.5 and 6.1 in \cite{applied-sde} for more details). Thus, simulating the entire forward process is not required. When the SDE is either Variance Preserving (VP), Variance Exploding, or sub-VP, its transition kernel takes the following functional form \cite{sde-score}:
\begin{equation*}
\label{eq:transition_form}
    p_{0t}(\matr{X}_t | \matr{X}_0) = \mathcal{N}\left( \matr{X}_t; m_t \matr{X}_0, \sigma_t^2 \matr{I} \right),
\end{equation*}
where $m_t$ is a scalar function of time, to which we refer to as the \emph{signal decay factor}. Similarly, the SDE for $\matr{A}_t$ has the same functional form.

The reverse diffusion of GDSS is defined by the system of SDEs:
\begin{equation}
\label{eq:system}
\footnotesize
\begin{cases}
    \dd{\matr{X}_t} = \left[ \matr{f}_{1,t}(\matr{X}_t) - g_{1,t}^2 \nabla_{\matr{X}_t} \log p_t(\matr{X}_t, \matr{A}_t) \right] \dd{\overline{t}} + g_{1,t} \dd{\overline{\matr{w}}_1} \\
    \dd{\matr{A}_t} = \left[ \matr{f}_{2,t}(\matr{A}_t) - g_{2,t}^2 \nabla_{\matr{A}_t} \log p_t(\matr{X}_t, \matr{A}_t) \right] \dd{\overline{t}} + g_{2,t} \dd{\overline{\matr{w}}_2}
\end{cases},
\end{equation}
where $\dd{\overline{\matr{w}_1}}$, $\dd{\overline{\matr{w}_2}}$ and $\dd{\overline{t}}$ correspond to the time reversal.

The partial scores $\nabla_{\matr{X}_t} \log p_t(\matr{X}_t, \matr{A}_t)$ and $\nabla_{\matr{A}_t} \log p_t(\matr{X}_t, \matr{A}_t)$ are modeled with neural networks $\matr{s}_{\theta, t}$ and $\matr{s}_{\phi, t}$, respectively, where $\theta$ and $\phi$ are the sets of parameters. The training objectives for these networks are the following:
\begin{align}
\footnotesize
\label{eq:objective}
\begin{split}
    \min_{\theta} \mathbb{E}_t \left\{ \lambda_1(t) \mathbb{E}_{\matr{G}_0} \mathbb{E}_{\matr{G}_t | \matr{G}_0} \| \matr{s}_{\theta, t} (\matr{G}_t) - \nabla_{\matr{X}_t} \log p_{0t} (\matr{X}_t | \matr{X}_0) \|^2_2 \right\} \\
    \min_{\phi} \mathbb{E}_t \left\{ \lambda_2(t) \mathbb{E}_{\matr{G}_0} \mathbb{E}_{\matr{G}_t | \matr{G}_0} \| \matr{s}_{\phi, t} (\matr{G}_t) - \nabla_{\matr{A}_t} \log p_{0t} (\matr{A}_t | \matr{A}_0) \|^2_2 \right\}
\end{split}
\end{align}
Positive weights $\lambda_1(t)$ and $\lambda_2(t)$ indicate the strength of score matching at time $t$.

\paragraph{The effect of hubs} In our setting, $\matr{G}_0$ denotes an uncorrupted ego-graph. Therefore, to estimate the expectation $\mathbb{E}_{\matr{G}_0}$  in Eq. \eqref{eq:objective}, we sample a random node $v$ from $\mathcal{G}$ and then build an ego-graph $\matr{G}_0 = \matr{G}_0(v)$. Since the score network for adjacency matrix $s_{\phi}(\matr{G}_t) \in \mathbb{R}^{N \times N}$, both training and inference scale quadratically w.r.t. the number of nodes. Hence, training GDSS might be computationally intractable for large graphs. Real-world networks may contain the so-called "hubs", i.e. nodes with a significant number of links \cite{scalingemergence, scalefree}. Consequently, even the 1-hop ego-graph of a hub can hinder mini-batch training. To alleviate this issue, we propose to sample subgraphs in which the number of nodes does not exceed some predefined value $M$. Specifically, if a sampled ego-graph has a number of nodes that exceeds $M$, we simply truncate it by subsampling a subgraph of size $M$. Note that in this subsampling procedure, we make sure that the subgraph contains the central node of the original ego-graph. For brevity, we do not differentiate between the original ego-graph and its subgraph.


\section{Ego-graph Reconstruction}
\label{sec:methods}

Let $\mathcal{G} = (\mathcal{V}, \mathcal{E}, \mathcal{X})$ be an unweighted attributed network with the vertex set $\mathcal{V}$, edge set $\mathcal{E}$, and node attribute matrix $\mathcal{X} \in \mathbb{R}^{|\mathcal{V}| \times F}$. The goal is to rank the nodes from $\mathcal{V}$ such that anomalous nodes are placed higher than normal nodes. A common strategy for solving node outlier detection is to design an unsupervised scoring function $\operatorname{score}(\cdot): \mathcal{V} \to \mathbb{R}$ that defines the node ranking \cite{gad-survey}. As a convention, we consider positive examples to be outliers. Consequently, larger scores should reflect higher degrees of abnormalities.

Each node $v$ of $\mathcal{G}$ induces a k-hop ego-graph $\matr{G}(v)$.  We assume that there is a hidden underlying distribution of ego-graphs $p_{\textrm{data}}$ from which the set of observed samples $\left\{ \matr{G}(v) \, | \, v \in \mathcal{V} \right\}$ is drawn. To solve ANOS ND, we propose to learn this distribution through GDSS \cite{sde-score}. We assign anomaly scores using a dissimilarity between the original and reconstructed ego-graphs with score-based generative models. Our key assumption is that the generator reconstructs inlier ego-graphs more accurately than outliers.

After learning the distribution of ego-graphs, we can calculate the reconstruction error for each node. Given node $v$ and time $\tau$, the reconstruction operator acts as follows:
\begin{enumerate}[nosep]
    \item Run the forward diffusion (Eq. \eqref{eq:forward}) on $\matr{G}(v)$ until time $\tau$, resulting in $\matr{G}_{\tau}(v)$
    \item Solve the reverse diffusion (Eq. \eqref{eq:system}) with an initial condition on $\matr{G}_{\tau}(v)$, resulting in $\hat{\matr{G}}(v, \tau)$ 
\end{enumerate}
We combine the ideas of sampling with different levels of noise \cite{ddpm-ood} and sampling multiple noisy versions of the same example \cite{ood-neighbor}. The ego-graphs are reconstructed several times with different values of $\tau=\tau_1, \dots, \tau_K$ distributed uniformly in $[0, T]$. At each time $\tau_i$, we sample $S$ noisy versions of the ego-graphs from $p_{0\tau_i}$.
Let us denote $\matr{G}^{(1)}_{\tau}(v), \matr{G}^{(2)}_{\tau}(v), \dots, \matr{G}^{(S)}_{\tau}(v)$ the independently corrupted ego-graphs centered around node $v$ at time $\tau$. Further, we denote $\hat{\matr{G}}^{(1)}(v, \tau), \hat{\matr{G}}^{(2)}(v, \tau), \dots, \hat{\matr{G}}^{(S)}(v, \tau)$ the corresponding reconstructions. The system in Eq. \eqref{eq:system} can be solved numerically with the Euler-Maruyama or Predictor-Corrector methods \cite{sde-score}. In addition, the authors of GDSS present a novel solver, Symmetric Splitting for System of SDEs (S4) \cite{sde-graph}.

Time $\tau$ is associated with the level of noise: the higher values of $\tau$ correspond to the lower values of the Signal-to-Noise Ratio (SNR). Given the variance of the perturbation kernel $\sigma^2_t$ and the signal decay factor $m_t$ (see Eq. \eqref{eq:transition_form}), SNR at time $\tau$ \cite{vdm} can be defined as
\begin{equation}
    \operatorname{SNR}(\tau) = \frac{m_{\tau}^2}{\sigma_{\tau}^2}.
\end{equation}
In the limit $\tau \to \infty$, the original signal is completely diminished, and the reconstruction operator acts blindly. Thus, we propose to reweight the errors at different noise scales by $\sqrt{\operatorname{SNR}(\tau)}$. 

Given a dissimilarity measure $d(\cdot, \cdot)$ on graphs of the same size and a time penalty function $\gamma(\cdot)$, we define the anomaly score for node $v$ as
\begin{equation}
\label{eq:anomaly_score}
    \operatorname{score}(v) = \sum_{i=1}^K \sum_{j=1}^S \left( \gamma(\tau_i) \cdot d(\matr{G}^{(j)}(v, \tau_i), \hat{\matr{G}}(v))\right).
\end{equation}
In this work, we set $\gamma(\tau)$ as either $\operatorname{SNR}(\tau)$ or $1$ (no weighting). As for ego-graph dissimilarity, we propose two different ways of measuring it: 1) as a convex combination of \textbf{matrix} distances; 2) as the difference in normalized \textbf{energies}. 

\paragraph{Matrix distance} One of the common ways to define a dissimilarity measure for graphs is to sum the distances between adjacency and feature matrices \cite{dominant}:
\begin{equation}
\label{eq:matrix_score}
    d(\matr{G}, \hat{\matr{G}}) = (1 - \alpha) \cdot \frac{\| \matr{A} - \hat{\matr{A}} \|_F}{N^2} + \alpha \cdot \frac{\| \matr{X} - \hat{\matr{X}} \|_F}{N \cdot F},
\end{equation}
where $\alpha \in [0, 1]$ is a hyperparameter, $N$ and $F$ are the numbers of nodes and features, respectively. In addition, we normalize matrices by their size. Normalizing the errors in adjacency matrices by their dimensionality helps to deal with the bias towards larger ego-graphs. Moreover, the normalization of the feature matrix error allows us to choose the weight $\alpha$ across different datasets more consistently since $F$ depends on the dataset.

\paragraph{Shift in energy} Let $\matr{D}$ be the diagonal matrix of node degrees and $\matr{L}$ be the normalized Laplacian:
\begin{equation}
    \matr{L} = \matr{D}^{\dagger/2} (\matr{D} - \matr{A}) \matr{D}^{\dagger/2}.
\end{equation}
Note that instead of taking the exact inverse square root of $\matr{D}$, we operate with its pseudoinverse square root. This allows us to normalize the Laplacian even if the reconstructed ego-graphs contain isolated nodes. If a graph is directed, then we symmetrize its Laplacian. Both features and structure can be incorporated into a single functional, the Dirichlet energy of a graph, which is defined as the variation of features along the edges \cite{note-oversmoothing}:
\begin{equation}
    \mathcal{E}(\matr{X}, \matr{L}) = \sum_{(i,j) \in E} \left\| \frac{\matr{X}_i}{\sqrt{D_{ii}}} - \frac{\matr{X}_j}{\sqrt{D_{jj}}} \right\|^2 = \Tr \matr{X}^{\intercal} \matr{L} \matr{X}.
\end{equation}
It can be interpreted as a measure of feature smoothness, with lower values of the energy indicating that the adjacent nodes have similar features. In general, the Dirichlet energy is unbounded above. Following \cite{gradient-flows}, we normalize the energy by the squared Frobenius norm of features, which yields a quantity bounded by the Laplacian spectral radius $\rho_\matr{L}$:
\begin{equation}
    0 \leq \frac{\mathcal{E}(\matr{X}, \matr{L})}{\|\matr{X}\|^2} \leq \rho_{\matr{L}} \leq 2.
\end{equation}
Contrary to the previous studies \cite{note-oversmoothing, dirichlet-constrained, gradient-flows}, we view energy as a functional of both features and the Laplacian. Bounding the energy helps quantify the shift in energy, which we define through the absolute difference between reconstructed and original energies:
\begin{equation}
\label{eq:shift}
    d(\matr{G}, \hat{\matr{G}}) = \left| \frac{\mathcal{E}(\matr{X}, \matr{L})}{\|\matr{X}\|^2} - \frac{\mathcal{E}(\hat{\matr{X}}, \hat{\matr{L}})}{\|\hat{\matr{X}}\|^2} \right|.
\end{equation}
A large gap between energies indicates a drastic change in how the node features align with each other as well as structural changes. Thus, a shift in energy can serve as a dissimilarity measure.


\section{Experiments}
\label{sec:experiments}

\begin{table*}[ht]
    \caption{ROC-AUC ($\%$) on datasets with organic outliers. The best average results are written in \textbf{bold}, and the best maximum results are \underline{underlined}. TLE and OOM\_C indicate that the method exceeded the time limit of 24 hours and failed to fit in VRAM, respectively.} 
    \label{tab:roc_metrics}
    \scriptsize
    \setlength\tabcolsep{4pt}
    \centering
    \vskip 0.1in
    \begin{tabular}{l|cccccc}
        \toprule
        \textbf{Algorithm} & \textbf{Weibo} & \textbf{Reddit} & \textbf{Disney} & \textbf{Books} & \textbf{Enron} & \textbf{DGraph} \\
        \midrule
        LOF & $56.5 \pm 0.0 \, (56.5)$ & $\mathbf{57.2} \pm 0.0 \, (57.2)$ & $47.9 \pm 0.0 \, (47.9)$ & $36.5 \pm 0.0 \, (36.5)$  & $46.4 \pm 0.0 \, (46.4)$ & TLE \\
        IF & $53.5 \pm 2.8 \, (57.5)$ & $45.2 \pm 1.7 \, (47.5)$ & $57.6 \pm 2.9 \, (63.1)$ & $43.0 \pm 1.8 \, (47.5)$ & $40.1 \pm 1.4 \, (43.1)$ & $\mathbf{60.9} \pm 0.7 (\underline{62.0})$ \\
        MLPAE & $82.1 \pm 3.6 \, (86.1)$ & $50.6 \pm 0.0 \, (50.6)$ & $49.2 \pm 5.7 (64.1)$ & $42.5 \pm 5.6 \, (52.6)$ & $73.1 \pm 0.0 \, (73.1)$ & $37.0 \pm 1.9 \, (41.3) $ \\
        \midrule
        SCAN & $63.7\pm 5.6 \, (70.8)$ & $ 49.9 \pm 0.3 \, (50.0)$ & $50.5 \pm 4.0 \, (56.1)$ & $49.8 \pm 1.7 \, (52.4)$ & $ 52.8 \pm 3.4 \, (58.1)$ & TLE \\
        Radar & $\mathbf{98.9} \pm 0.1 (\underline{99.0})$ & $54.9 \pm 1.2 \, (56.9)$ & $51.8 \pm 0.0 \, (51.8)$ & $52.8 \pm 0.0 \, (52.8)$ & $\mathbf{80.8} \pm 0.0 \, (80.8)$ & OOM\_C \\
        ANOMALOUS & $\mathbf{98.9} \pm 0.1 \, (\underline{99.0})$ & $54.9 \pm 5.6 \, (\underline{60.4})$ & $51.8 \pm 0.0 \, (51.8)$ & $52.8 \pm 0.0 \, (52.8)$ & $\mathbf{80.8} \pm 0.0 \, (80.8)$ & OOM\_C \\
        \midrule
        GCNAE & $90.8 \pm 1.2 \, (92.5)$ & $50.6 \pm 0.0 \, (50.6)$ & $42.2 \pm 7.9 \, (52.7)$ & $50.0 \pm 4.5 \, (57.9)$ & $66.6 \pm 7.8 \, (80.1)$ & $40.9 \pm 0.5 \, (42.2)$ \\
        DOMINANT & $85.0 \pm 14.6 \, (92.5)$ & $ 56.0 \pm 0.2 \, (56.4)$ & $47.1 \pm 4.5 \, (54.9)$ & $ 50.1 \pm 5.0 \, (58.1)$ & $73.1 \pm 8.9 \, (\underline{85.0})$ & OOM\_C \\
        DONE & $85.3 \pm 4.1 \, (88.7)$ & $53.9 \pm 2.9 \, (59.7)$ & $41.7 \pm 6.2 \, (50.6)$ & $43.2 \pm 4.0 \, (52.6)$ & $46.7\pm 6.1 \, (67.1)$ & OOM\_C \\
        AdONE & $84.6 \pm 2.2 \, (87.6)$ & $50.4 \pm 4.5 \, (58.1)$ & $48.8 \pm 5.1 \, (59.2)$ & $53.6 \pm 2.0 \, (56.1)$ & $44.5 \pm 2.9 \, (53.6)$ & OOM\_C \\
        AnomalyDAE & $91.5 \pm 1.2 \, (92.8)$ & $55.7 \pm 0.4 \, (56.3)$ & $48.8 \pm 2.2 \, (55.4)$ & $\mathbf{62.2} \pm 8.1 \, (\underline{73.2})$ & $54.3 \pm 11.2 \, (69.1)$ & OOM\_C \\
        GAAN & $92.5 \pm 0.0 \, (92.5)$ & $ 55.4 \pm 0.4 \, (56.0)$ & $48.0 \pm 0.0 \, (48.0)$ & $ 54.9 \pm 5.0 \, (61.9)$ & $73.1 \pm 0.0 \, (73.1)$ & OOM\_C \\
        GUIDE & OOM\_C & OOM\_C & $38.8 \pm 8.9 \, (52.5)$ & $48.4 \pm 4.6 \, (63.5)$ & OOM\_C & OOM\_C \\
        CONAD & $85.4 \pm 14.3 \, (92.7)$ & $56.1 \pm 0.1 \, (56.4)$ & $48.0 \pm 3.5 \, (53.1)$ & $52.2 \pm 6.9 \, (62.9)$ & $71.9 \pm 4.9 \, (84.9)$ & $34.7 \pm 1.2 \, (36.5)$ \\
        \midrule
        \textbf{Rec} & $74.5 \pm 12.6 \, (88.6)$ & $44.4 \pm 0.4 \, (45.1)$ & $\mathbf{65.0} \pm 11.2 \, (79.8)$ & $56.9 \pm 2.7 \, (62.1)$ & $44.0 \pm 4.0 \, (51.4)$ & TLE \\
        \textbf{Rec (unweighted)} & $74.4 \pm 12.4 \, (88.2)$ & $44.5 \pm 0.4 \, (45.2)$ & $63.1 \pm 12.9 \, (78.1)$ & $57.1 \pm 2.8 \, (62.7)$ & $44.0 \pm 4.2 \, (51.4)$ & TLE \\
        \textbf{Energy} & $51.3 \pm 11.0 \, (64.8)$ & $55.1 \pm 0.8 \, (56.7)$ & $56.7 \pm 5.7 \, (67.0)$ & $52.4 \pm 3.8 \, (59.2)$ & $36.5 \pm 5.6 \, (48.2)$ & TLE \\
        \textbf{Energy (unweighted)} & $51.9 \pm 11.1 \, (67.8)$ & $55.0 \pm 0.9 \, (56.8)$ & $58.4 \pm 5.9 \, (68.2)$ & $52.7 \pm 3.0 \, (57.5)$ & $35.5 \pm 5.0 \, (46.9)$ & TLE \\
        \bottomrule
    \end{tabular}
    \vskip -0.1in
\end{table*}

To assess the quality of our method, we follow the evaluation protocol from the BOND benchmark \cite{bond}. This benchmark tests 14 different approaches, ranging from matrix factorization to deep neural networks. For fair comparison, tuning of hyperparameters is performed on the shared grid. BOND evaluates the algorithms on real-world networks that include organic and synthetic anomalies. In this work, we evaluate our methods only on graphs with organic outliers. A detailed description of datasets as well as their statistics is provided in Appendix \ref{appendix:datasets}. In the preprocessing step, we standardize the node feature matrices such that each feature has a unit standard deviation. This allows us to use the same level of noise for each feature during forward diffusion.

If possible, the models from the BOND benchmark share the same grid of hyperparameters. On each dataset, performance is evaluated $20$ times. At the beginning of a trial, a model is built with randomly drawn hyperparameters. To solve the reverse diffusion SDEs, we use the Euler-Maruyama solver with $\lfloor 100 \cdot \frac{\tau}{T} \rfloor$ steps. We set the number of reconstruction levels $K=4$ and the number of samples per level $S=3$. More details regarding the architecture and hyperparameters can be found in Appendix \ref{appendix:implementation}. In Appendix \ref{appendix:solver}, we motivate our choice of SDE solver.

Tables \ref{tab:roc_metrics}, \ref{tab:ap_metrics}, and \ref{tab:recall_metrics} show a comparison of our methods and the baselines in terms of metrics (see Appendix \ref{appendix:metrics} for the description of metrics and additional results). The baseline results are taken from BOND. Our results are shown in the bottom rows (in \textbf{bold}). \emph{Rec} stands for reconstruction-based detection with matrix distance, whereas \emph{Energy} corresponds to setting a shift in energy as a dissimilarity measure. By default, we assume that the scores are weighted with the SNR time penalties. We first write the average metrics, the standard deviation, and the maximum in brackets. Ego-graphs are visualized in Appendix \ref{appendix:graphs}.

\paragraph{Energy reconstruction} Further, we investigate how well GDSS reconstructs the energy. Figure \ref{fig:energy} shows the original and reconstructed normalized energies from different noise levels. The energy values are collected using all 20 checkpoints. As can be seen from the histograms, GDSS tends to generate ego-graphs with low energies. This effect is more visible in Figure \ref{fig:diff_energy}, where we plotted histograms of the signed shift in energy $\frac{\mathcal{E}(\matr{X}, \matr{L})}{\|\matr{X}\|^2} - \frac{\mathcal{E}(\hat{\matr{X}}, \hat{\matr{L}})}{\|\hat{\matr{X}}\|^2}$. A significant bias towards positive values indicates that the reconstructed ego-graphs are either smoother or sparser than their originals.


\section{Discussion}
\label{sec:discussion}
In this work, we present a novel method to tackle node outlier detection in attributed networks. Our approach leverages score-based generative graph models to reconstruct ego-graphs and is agnostic to the particular choice of model. We assign anomaly scores based on a dissimilarity measure between the original and reconstructed ego-graphs. We experiment with two ways of measuring the dissimilarity: 1) combining the norms of the differences between both the adjacency and feature matrices; and 2) calculating the absolute shift in normalized Dirichlet energies. The former measure shows the best results on Disney dataset, whereas the rest of the benchmarked methods completely fail. However, it shows results that are poor on two larger datasets, Reddit and Enron, and moderate on the others. The latter measure is consistently worse than the matrix distance, except on Reddit, whose energy distribution forms a narrow band located on smaller values, as opposed to the other graphs. Analyzing the shift in energy might be helpful not only for anomaly detection in networks but also for assessing the quality of graph generative models. 

Future directions involve finding an optimal architecture and incorporating more expressive graph convolutions, such as GRAFF \cite{gradient-flows}. Another prospect is guided generation with node positional encodings. Unfolding a graph into a collection of ego-graphs comes with both advantages and shortcomings. It serves as a technique for training score-based models at a local scale, efficiently learning the notion of normality in interactions. Nevertheless, information from neighborhoods may not be sufficient. Employing positional encodings might alleviate this issue by capturing subtle higher-order interactions. They include, but are not limited to, Laplacian PE \cite{laplacian-pe}, SignNet \cite{signnet}, and geodesics \cite{graphgps}.


\section*{Acknowledgements}
This work was supported by Ministry of Science and Higher Education grant No. 075-10-2021-068.


\bibliography{example_paper}
\bibliographystyle{icml2023}

\newpage
\appendix
\onecolumn


\section{Related Work}
\label{appendix:review}

\subsection{Anomaly Detection}
In \cite{gad-survey}, the authors present a survey on graph anomaly detection methods. For ANOS ND, they provide the following taxonomy of anomalies: global, in which node attributes differ from the distribution of attributes; structural, in which the graph's structure is considered; and community anomalies, in which attributes differ within the nodes in the same community. They also review work that assigns anomaly scores to individual nodes based on their contribution to the network reconstruction error. The BOND benchmark \cite{bond} investigates the performance of
\begin{itemize}[nosep]
    \item graph-agnostic algorithms: LOF \cite{lof}, IF \cite{isolationbased}, MLPAE \cite{mlpae}
    \item classical algorithms: SCAN \cite{scan}, RADAR \cite{radar}, ANOMALOUS \cite{anomalous}
    \item deep algorithms: GCNAE \cite{gcnae}, DOMINANT \cite{dominant}, DONE, AdONE \cite{done}, GAAN \cite{gaan}, GUIDE \cite{guide}, CONAD \cite{conad}.
\end{itemize}

For instance, DOMINANT \cite{dominant} maps nodes of a graph into the latent space through a graph convolutional encoder \cite{gcn}. Then, the latent representations are passed into two separate decoders that reconstruct the adjacency matrix and node features, respectively. The nodes are scored with the weighted sum of the corresponding reconstruction errors.

\paragraph{Ego-networks} Previous work regarding the analysis of ego-networks includes the non-deep learning algorithms OddBall \cite{oddball} and AMEN \cite{amen}. The former solves ANOS ND on weighted unattributed graphs by employing outlier detection on the graph statistics of ego-networks. OddBall relies on a set of heavy assumptions as to what can be considered an anomaly, e.g., star-shaped or near-clique ego-networks. Therefore, it might not generalize well to arbitrary graphs. Further, AMEN introduces the normality score for attributed subgraphs. For a given subgraph, it measures how its nodes are similar to each other on some subset of attributes, as well as how they are dissimilar to the nodes from the boundary (a set of nodes that do not belong to the subgraph but have at least one neighbor from it).

\subsection{Score-Based Modeling for OOD detection}

The idea of using diffusion-based generative models in OOD detection appears in \cite{ood-neighbor, ddpm-ood}. Particularly, they consider the problem of OOD detection in the image domain. In \cite{ood-neighbor}, the authors propose to sample neighbors in the input space for a given image by generating them via a diffusion model, which is pre-trained on the in-distribution samples. The neighborhood is constructed by sampling the corrupted images from the forward transition distribution $q(\matr{x}_t | \matr{x}_0)$. A hyperparameter $t \in [0, T]$ defines the level of corruption. Then, the  neighborhood is passed to a discriminator to extract features (e.g., a ResNet \cite{resnet} pre-trained on an image classification task for in-distribution data). Finally, they compute the OOD scores as the sum of absolute differences between the features of a given image and its neighbors. 

An alternative approach, proposed in \cite{ddpm-ood}, consists of sampling the corrupted images with different levels of noise, distributed uniformly. The OOD scores are assigned as the combination of the MSE in image space and the LPIPS (Learned Perceptual Image Patch Similarity \cite{lpips}) in AlexNet's feature space \cite{alexnet}.

\subsection{Graph Generation}
As for graph generative models, there is a great variety of deep-learning methods. They include sequential modeling through variational auto-encoders \cite{graphvae}, recurrent neural networks \cite{graphrnn}, normalizing flows \cite{gnf} and score-based models \cite{scorebased, sde-graph}. However, most of the methods consider only non-attributed graphs. 

The idea of leveraging score-based models to generate graphs is first explored in \cite{scorebased}. The authors propose the process of denoising the adjacency matrix. They inject the permutation invariance of the underlying distribution of adjacency matrices as an inductive bias. It is done by designing a permutation equivariant score network, which can be constructed of message passing layers. In particular, the authors introduce EDP-GNN, a graph neural network architecture that is inspired by image dense prediction networks. It consists of the GNN layers that operate on multi-channel adjacency matrices, which are analogous to the feature maps produced by image convolutional layers. 

Further, in \cite{sde-graph}, the authors introduce Graph Diffusion via the System of Stochastic differential equations (GDSS) as an extension of the previous method to attributed graphs. They consider continuous-time diffusion, where both node features and the adjacency matrix are perturbed simultaneously. Forward and reverse diffusion processes are induced by stochastic differential equations (SDEs). Solving the reverse diffusion is especially expensive due to the high dimensionality of the score function. Hence, the authors instead aim to approximate the partial score functions that correspond to the node features and adjacency matrix. This results in a system of SDEs that is equivalent to the original reverse diffusion SDE. To that end, GDSS shows state-of-the-art performance in generating generic graphs and competitive performance in molecule generation.

Another diffusion-based model for graph generation, DiGress \cite{digress}, considers node and edge features to be discrete and drawn from the categorical distribution. Thus, each node and edge is attributed with exactly one type from the attribute space. The structural information is stored in the edge features by introducing the absence of an edge as a separate edge type. Graphs are perturbed through random transformations of their attributes. The transition matrices describe the probabilities of jumping from one type to another. To sample graphs, one needs to design a tractable prior distribution. A naive approach would be to choose uniform distribution over the attribute types as the prior. However, a noisy graph from this prior is highly dense, for which many denoising steps are required. Instead, the authors propose to use the product of the marginal distributions of node and edge types. As a result, the terminal graphs from the prior are much closer to the original ones. DiGress can also be applied to a case of continuous Gaussian noise. Such a modification, ConGress, is similar to GDSS yet models the full score function.


\section{Datasets}
\label{appendix:datasets}
\begin{itemize}
    \item \textbf{Weibo} \cite{weibo} is the users-posts-hashtags graph from the social platform of the same name. The users are considered suspicious (anomalies) if their posting frequency resembles that of bots (i.e., every $x$ seconds). BOND employs the \textbf{directed} user-user form of the same graph, with hashtags serving as the edges. The node attributes include aggregated information from users' posts, such as the post's location and the text's bag-of-word vectors. Further, the attributes are compressed via dimensionality reduction techniques. Since anomaly labels are derived from timestamps, temporal information is removed from the feature space.
    \item \textbf{Reddit} \cite{reddit-temporal, reddit-decoupling} is a subset of the user-group (user-subreddit) bipartite graph of the corresponding social media platform. Although both users and groups are considered nodes, there is no label to differentiate between the two. The banned users are assumed to be outliers. The LIWC \cite{liwc} representations of posts are aggregated for each node to construct features.
    \item \textbf{Disney} \cite{stat-mining} is a co-purchase network of movies from Amazon \cite{amazon-dynamics} with manually labeled anomalies. Each movie is attributed with its price, rating, number of reviews, etc.
    \item \textbf{Books} \cite{stat-mining} originates from the Amazon co-purchase network \cite{amazon-dynamics}, similar to the Disney dataset. The items labeled with the \emph{amazonfail} tag are considered outliers.
    \item \textbf{Enron} \cite{stat-mining} is an email communication network, where nodes are email addresses and edges are messages. Spam senders are labeled as outliers \cite{enron-spam}. Each email address is described by statistics such as the average message length and the average number of recipients.
    \item \textbf{DGraph} \cite{dgraph} is a financial social network where the nodes represent user accounts. The edge between two users exists if one of them adds the other as an emergency contact. Users with an overdue history are regarded as anomalous. The features include general information such as age, gender, and repayment dates.
\end{itemize}

\begin{table}[h]
    \caption{The statistics of datasets from BOND. \textbf{Ratio} indicates the ratios of outliers in a graph.}
    \centering
    \vskip 0.15in
    \begin{tabular}{l|ccccc}
        \toprule
        \textbf{Dataset} & \textbf{\#Nodes} & \textbf{\#Edges} & \textbf{\#Features} & \textbf{Avg. Degree} & \textbf{Ratio} \\
        \midrule
         \textbf{Weibo} & 8,405 & 407,963 & 400 & 48.5 & 10.3\% \\
         \textbf{Reddit} & 10,984 & 168,016 & 64 & 15.3 & 3.3\% \\
         \textbf{Disney} & 124 & 335 & 28 & 2.7  & 4.8\% \\
         \textbf{Books} & 1,418 & 3,695 & 21 & 2.6 & 2.0\% \\
         \textbf{Enron} & 13,533 & 176,987 & 18 & 13.1 & 0.4\% \\
         \textbf{DGraph} & 3,700,550 & 4,300,999 & 17 & 1.2 & 0.4\% \\
        \bottomrule
    \end{tabular}
    \vskip -0.1in
    \label{tab:bond_statistics}
\end{table}


\section{Implementation Details}
\label{appendix:implementation}

\paragraph{Architecture} To speed up the evaluation, we use a lightweight variant of GDSS. For score networks $s_{\theta, t}$ and $s_{\phi, t}$, we set the number of GCN and GMH layers to $1$. The number of heads for GMH is set to $4$. The inputs to GMH are the node feature matrix $\matr{X}$ and the adjacency tensor $[\matr{A}, \matr{A}^2]$. The output of GMH has four channels. The MLP inside the GMH block has two linear layers. GCN and GMH are followed by the channel-mixing MLPs that have three layers. All MLP blocks have the ELU activation \cite{elu}. We set the form of the forward diffusion SDEs to be Variance Preserving (VP SDE):
\begin{equation}
    \dd{\matr{G}_t} = -\frac12 \beta(t) \matr{G}_t \dd{t} + \sqrt{\beta(t)} \dd{\matr{w}_t},
\end{equation}
where $\beta(t) = \beta_{\min} + (\beta_{\max} - \beta_{\min}) t$, $\beta_{\min} = 0.1$, $\beta_{\max} = 1$.

\paragraph{Hyperparameters} Table \ref{tab:hyperparameters} presents the pool of hyperparameters common to the algorithms from BOND. \emph{Alpha} denotes the balancing weight for reconstructing the structure and features. The deep learning models are optimized with the Adam algorithm \cite{adam}. Note that Table \ref{tab:hyperparameters} does not include batch sizes. Due to the high memory consumption of our methods, we set them independently of the BOND benchmark. The corresponding batch sizes for each dataset are shown in Table \ref{tab:batch_size}.

\begin{table}[h]
    \caption{The grid of hyperparameters shared by the algorithms from BOND.}
    \label{tab:hyperparameters}
    \centering
    \vskip 0.15in
    \begin{tabular}{c|c|c|c|c|c|c}
        \toprule
        \textbf{Hyperparameters} & \textbf{Weibo} & \textbf{Disney} & \textbf{Books} & \textbf{Enron} & \textbf{DGraph} & \textbf{Reddit} \\
        \midrule
        learning rate & \multicolumn{6}{c}{$[0.1, 0.05, 0.01]$} \\
        \hline
        weight decay & \multicolumn{6}{c}{$0.01$} \\
        \hline
        epoch & \multicolumn{4}{c|}{$300$} & $2$ & $300$ \\
        \hline
        alpha & \multicolumn{6}{c}{$[0.8, 0.5, 0.2]$} \\
        \hline
        hid. dim. & $[32, 64, 128, 256]$ & \multicolumn{4}{c|}{$[8, 12, 16]$} & $[32, 48, 64]$ \\ 
        \bottomrule 
    \end{tabular}
    \vskip -0.1in
\end{table}

\begin{table}[h]
    \caption{Batch sizes used for training GDSS and inference.}
    \label{tab:batch_size}
    \centering
    \vskip 0.15in
    \begin{tabular}{c|c|c|c|c|c}
        \toprule
        \textbf{Weibo} & \textbf{Disney} & \textbf{Books} & \textbf{Enron} & \textbf{DGraph} & \textbf{Reddit} \\
        \midrule
        2048 & full batch & full batch & $4096$ & - & $4096$ \\
        \bottomrule
    \end{tabular}
    \vskip -0.1in
\end{table}

\paragraph{Software} All methods are written in Python 3 and use PyTorch for autodifferentiation \cite{pytorch}. We employ both DGL (Deep Graph Library \cite{dgl} and PyG (PyTorch Geometric \cite{pyg} for graphs, which are popular libraries for training GNNs. For evaluation, we use the code from PyGOD, a library for graph outlier detection \cite{pygod}. Our code and model checkpoints are publicly available at \url{https://github.com/realfolkcode/GraphDiffusionAnomaly}.


\section{Metrics}
\label{appendix:metrics}

\begin{itemize}
    \item \textbf{ROC-AUC} assesses the quality of predicted scores by taking into account all possible thresholds that separate negative and positive examples. At each threshold value, the true positive rate (TPR) and the false positive rate (FPR) are calculated. Then, the Receiver Operating Curve is formed by plotting (FPR, TPR) pairs. ROC-AUC is an integral measure defined as the area under the ROC curve. One of the popular interpretations is the probability of a random positive example (anomaly) having a higher score than a random negative example. ROC-AUC equals $1$ means that the algorithm perfectly separates anomalies from normal nodes, whereas ROC-AUC equals $0.5$ indicates that the model makes random guesses.
    \item \textbf{Average Precision} is calculated as follows:
    \begin{equation}
        \operatorname{AP} = \sum_{n}\left( R_n - R_{n-1} \right) P_n,
    \end{equation}
    where $n$ is the threshold index, $P_n$ and $R_n$ are the precision and recall at the $n$-th threshold, respectively \cite{sklearn}. It can be seen as a summarization of the precision-recall curve.
    \item \textbf{Recall@k} indicates the fraction of true outliers among the top-$k$ ranked examples. In the BOND benchmark, $k$ is set as the number of anomalies in a dataset. Hence, Recall@k measures how well the model places the outliers at the top of the list.
\end{itemize}

\begin{table}[h]
    \caption{Average Precision ($\%$) on datasets with organic outliers. The best average results are written in \textbf{bold}, and the best maximum results are \underline{underlined}. TLE and OOM\_C indicate that the method exceeded the time limit of 24 hours and failed to fit in VRAM, respectively.}
    \label{tab:ap_metrics}
    \scriptsize
    \setlength\tabcolsep{4pt}
    \centering
    \vskip 0.15in
    \begin{tabular}{l|cccccc}
        \toprule
        \textbf{Algorithm} & \textbf{Weibo} & \textbf{Reddit} & \textbf{Disney} & \textbf{Books} & \textbf{Enron} & \textbf{DGraph} \\
        \midrule
        LOF & $15.8 \pm 0.0 \, (15.8)$ & $\mathbf{4.2} \pm 0.0 \, (4.2)$ & $5.2 \pm 0.0 \, (5.2)$ & $1.5 \pm 0.0 \, (1.5)$ & $0.0 \pm 0.0 \, (0.0)$ & TLE \\
        IF & $12.9 \pm 2.6 \, (19.8)$ & $2.8 \pm 0.1 \, (2.9)$ & $10.1 \pm 4.5 \, (22.6)$ & $ 1.9 \pm 0.2 \, (2.7)$ & $0.1 \pm 0.0 \, (0.1)$ & $\mathbf{1.8} \pm 0.0 \, (\underline{1.9})$ \\
        MLPAE & $52.8\pm 9.9 \, (64.5)$ & $3.4 \pm 0.0 \, (3.4)$ & $5.9 \pm 0.8 \, (7.9)$ & $1.8 \pm 0.3 \, (2.5)$ & $0.1 \pm 0.0 \, (0.1)$ & $0.9 \pm 0.0 \, (1.0)$ \\
        \midrule
        SCAN & $17.3 \pm 3.4 \, (20.5)$ & $3.3 \pm 0.0 \, (3.3)$ & $5.0 \pm 0.3 \, (5.5)$ & $2.0 \pm 0.1 \, (2.1)$ & $0.0 \pm 0.0 \, (0.1)$ & TLE \\
        Radar & $\mathbf{92.1} \pm 0.7 \, (\underline{92.9})$ & $3.6 \pm 0.2 \, (3.9)$ & $7.2 \pm 0.0 \, (7.2)$ & $2.2 \pm 0.0 \, (2.2)$ & $\mathbf{0.2} \pm 0.0 \, (0.2)$ & OOM\_C \\
        ANOMALOUS & $\mathbf{92.1} \pm 0.7 \, (\underline{92.9})$ & $4.0 \pm 0.6 \, (\underline{5.1})$ & $7.2 \pm 0.0 \, (7.2)$ & $2.2 \pm 0.0 \, (2.2)$ & $\mathbf{0.2} \pm 0.0 \, (0.2)$ & OOM\_C \\
        \midrule
        GCNAE & $70.8 \pm 5.0 \, (80.9)$ & $3.4 \pm 0.0 \, (3.4)$ & $4.8 \pm 0.7 \, (5.8)$ & $2.1 \pm 0.4 \, (3.5)$ & $0.1 \pm 0.0 \, (0.1)$ & $1.0 \pm 0.0 \, (1.0)$ \\
        DOMINANT & $18.0 \pm 10.2 \, (36.2)$ & $3.7 \pm 0.0 \, (3.8)$ & $7.6 \pm 5.0 \, (23.2)$ & $2.2 \pm 0.6 \, (4.1)$ & $0.1 \pm 0.1 \, (\underline{0.4})$ & OOM\_C \\
        DONE & $65.5 \pm 13.4 \, (77.3)$ & $3.7 \pm 0.4 \, (4.5)$ & $5.0 \pm 0.7 \, (6.4)$ & $1.8 \pm 0.3 \, (2.6)$ & $0.1 \pm 0.0 \, (0.1)$ & OOM\_C \\
        AdONE & $62.9 \pm 9.5 \, (74.4)$ & $3.3 \pm 0.4 \, (4.0)$ & $6.1 \pm 1.5 \, (11.7)$ & $2.5 \pm 0.3 \, (3.2)$ & $0.1 \pm 0.0 \, (0.1)$ & OOM\_C \\
        AnomalyDAE & $38.5 \pm 22.5 \, (77.3)$ & $3.7 \pm 0.1 \, (3.8)$  & $5.7 \pm 0.2 \, (6.3)$ & $\mathbf{3.5} \pm 1.4 \, (\underline{7.8})$ & $0.1 \pm 0.0 \, (0.1)$ & OOM\_C \\
        GAAN & $80.3 \pm 0.2 \, (80.7)$ & $3.7 \pm 0.1 \, (3.9)$ & $5.6 \pm 0.0 \, (5.6)$ & $2.6 \pm 0.8 \, (5.6)$ & $0.1 \pm 0.0 \, (0.1)$ & OOM\_C \\
        GUIDE & OOM\_C & OOM\_C & $4.8 \pm 0.9 \, (6.9)$ & $1.9 \pm 0.3 \, (3.1)$ & OOM\_C & OOM\_C \\
        CONAD & $15.6 \pm 6.9 \, (31.7)$ & $3.7 \pm 0.3 \, (4.6)$ & $6.0 \pm 1.4 \, (11.5)$ & $2.5 \pm 0.8 \, (4.9)$ & $0.1 \pm 0.0 \, (0.3)$ & $0.9 \pm 0.0 \, (0.9)$ \\
        \midrule
        \textbf{Rec} & $28.6 \pm 9.7 \, (42.4)$ & $2.9 \pm 0.1 \, (3.2)$ & $13.9 \pm 6.5 \, (33.6)$ & $2.9 \pm 0.8 \, (6.4)$ & $0.1 \pm 0.0 \, (0.1)$ & TLE \\
        \textbf{Rec (unweighted)} & $29.2 \pm 10.5 \, (43.0)$ & $2.9 \pm 0.1 \, (3.2)$ & $14.6 \pm 7.0 \, (30.8)$ & $2.8 \pm 0.6 \, (4.4)$ & $0.1 \pm 0.0 \, (0.2)$ & TLE \\
        \textbf{Energy} & $10.7 \pm 2.4 \, (14.1)$ & $4.0 \pm 0.4 \, (5.0)$ & $15.9 \pm 7.4 \, (29.2)$ & $2.7 \pm 0.7 \, (4.4)$ & $0.0 \pm 0.0 \, (0.1)$ & TLE \\
        \textbf{Energy (unweighted)} & $11.0 \pm 2.9 \, (18.0)$ & $3.9 \pm 0.3 \, (5.0)$ & $\mathbf{17.2} \pm 8.0 \, (29.0)$ & $2.6 \pm 0.6 \, (3.8)$ & $0.0 \pm 0.0 \, (0.0)$ & TLE \\
        \bottomrule
    \end{tabular}
    \vskip -0.1in
\end{table}

\begin{table}[h]
    \caption{Recall@k ($\%$) on datasets with organic outliers. The best average results are written in \textbf{bold}, and the best maximum results are \underline{underlined}. TLE and OOM\_C indicate that the method exceeded the time limit of 24 hours and failed to fit in VRAM, respectively.}
    \label{tab:recall_metrics}
    \scriptsize
    \setlength\tabcolsep{4pt}
    \centering
    \vskip 0.15in
    \begin{tabular}{l|cccccc}
        \toprule
        \textbf{Algorithm} & \textbf{Weibo} & \textbf{Reddit} & \textbf{Disney} & \textbf{Books} & \textbf{Enron} & \textbf{DGraph} \\
        \midrule
        LOF & $22.0 \pm 0.0 \, (22.0)$ & $\mathbf{4.4} \pm 0.0 \, (4.4)$ & $0.0 \pm 0.0 \, (0.0)$ & $0.0 \pm 0.0 \, (0.0)$ & $0.0 \pm 0.0 \, (0.0)$ & TLE \\
        IF & $13.8 \pm 6.4 \, (24.3)$ & $0.1 \pm 0.1 \, (0.3)$ & $9.2 \pm 8.3 \, (16.7)$ & $1.1 \pm 1.6 \, (3.6)$ & $0.0 \pm 0.0 \, (0.0)$ & $0.1 \pm 0.1 \, (0.4)$ \\
        MLPAE & $48.9 \pm 11.0 \, (62.1)$ & $3.0 \pm 0.0 \, (3.0)$ & $0.0 \pm 0.0 \, (0.0)$ & $0.9 \pm 1.6 \, (3.6)$ & $0.0 \pm 0.0 \, (0.0)$ & $\mathbf{0.5} \pm 0.1 \, (\underline{0.6})$ \\
        \midrule
        SCAN & $23.8 \pm 7.0 \, (30.5)$ & $2.7 \pm 0.3 \, (3.0)$ & $7.5 \pm 11.2 \, (\underline{33.3})$ & $0.7 \pm 1.4 \, (3.6)$ & $0.0 \pm 0.0 \, (0.0)$ & TLE \\
        Radar & $\mathbf{86.4} \pm 0.8 \, (\underline{87.4})$ & $2.1 \pm 0.8 \, (3.5)$ & $0.0 \pm 0.0 \, (0.0)$ & $0.0 \pm 0.0 \, (0.0)$ & $0.0 \pm 0.0 \, (0.0)$ & OOM\_C \\
        ANOMALOUS & $\mathbf{86.4} \pm 0.8 \, (\underline{87.4})$ & $4.0 \pm 1.9 \, (\underline{7.9})$ & $0.0 \pm 0.0 \, (0.0)$ & $0.0 \pm 0.0 \, (0.0)$ & $0.0 \pm 0.0 \, (0.0)$ & OOM\_C \\
        \midrule
        GCNAE & $67.6 \pm 5.2 \, (77.3)$ & $3.0 \pm 0.0 \, (3.0)$ & $0.0 \pm 0.0 \, (0.0)$ & $0.7 \pm 1.8 \, (7.1)$ & $0.0 \pm 0.0 \, (0.0)$ & $0.4 \pm 0.0 \, (0.4)$ \\
        DOMINANT & $19.7 \pm 13.8 \, (37.4)$ & $0.9 \pm 0.4 \, (2.7)$ & $3.3 \pm 6.7 \, (16.7)$ & $1.6 \pm 3.1 \, (\underline{10.7})$ & $0.0 \pm 0.0 \, (0.0)$ & OOM\_C \\
        DONE & $65.4 \pm 12.4 \, (76.3)$ & $2.8 \pm 1.6 \, (5.7)$ & $0.0 \pm 0.0 \, (0.0)$ & $1.1 \pm 1.6 \, (3.6)$ & $0.0 \pm 0.0 \, (0.0)$ & OOM\_C \\
        AdONE & $64.3 \pm 7.6 \, (74.3)$ & $1.0 \pm 1.2 \, (3.8)$ & $1.7 \pm 5.0 \, (16.7)$ & $\mathbf{3.0} \pm 1.7 \, (7.1)$ & $0.0 \pm 0.0 \, (0.0)$ & OOM\_C \\
        AnomalyDAE & $42.2 \pm 23.7 \, (75.7)$ & $0.9 \pm 0.5 \, (3.0)$ & $ 0.0 \pm 0.0 \, (0.0)$ & $2.7 \pm 2.2 \, (7.1)$ & $0.0 \pm 0.0 \, (0.0)$ & OOM\_C \\
        GAAN & $77.1 \pm 0.2 \, (77.4)$ & $1.1 \pm 0.4 \, (2.2)$ & $0.0 \pm 0.0 \, (0.0)$ & $1.8 \pm 1.8 \, (3.6)$ & $0.0 \pm 0.0 \, (0.0)$ & OOM\_C \\
        GUIDE & OOM\_c & OOM\_C & $0.0 \pm 0.0 \, (0.0)$ & $0.4 \pm 1.1 \, (3.6)$ & OOM\_C & OOM\_C \\
        CONAD & $20.3 \pm 13.3 \, (37.1)$ & $1.3 \pm 1.6 \, (7.6)$ & $0.8 \pm 3.6 \, (16.7)$ & $1.7 \pm 2.9 \, (\underline{10.7})$ & $0.0 \pm 0.0 \, (0.0)$ & $0.4 \pm 0.1 \, (\underline{0.6})$ \\
        \midrule
        \textbf{Rec} & $35.4 \pm 12.9 \, (50.8)$ & $2.2 \pm 0.8 \, (4.4)$ & $13.6 \pm 11.2 \, (\underline{33.3})$ & $2.1 \pm 2.6 \, (7.1)$ & $0.0 \pm 0.0 \, (0.0)$ & TLE \\
        \textbf{Rec (unweighted)} & $36.4 \pm 14.5 \, (52.7)$ & $2.4 \pm 0.7 \, (4.1)$ & $14.2 \pm 12.5 \, (\underline{33.3})$ & $2.4 \pm 2.8 \, (\underline{10.7})$ & $0.0 \pm 0.0 \, (0.0)$ & TLE \\
        \textbf{Energy} & $8.8 \pm 4.3 \, (17.6)$ & $4.2 \pm 1.4 \, (7.1)$ & $13.3 \pm 10.0 \, (\underline{33.3})$ & $2.7 \pm 3.2 \, (\underline{10.7})$ & $0.0 \pm 0.0 \, (0.0)$ & TLE \\
        \textbf{Energy (unweighted)} & $9.2 \pm 5.0 \, (24.2)$ & $3.6 \pm 1.1 \, (6.8)$ & $\mathbf{15.8} \pm 12.3 \, (\underline{33.3})$ & $2.5 \pm 2.8 \, (7.1)$ & $0.0 \pm 0.0 \, (0.0)$ & TLE \\
        \bottomrule
    \end{tabular}
    \vskip -0.1in
\end{table}


\section{Solver Comparison}
\label{appendix:solver}

For the reconstruction approach, we motivate our choice of solver (Euler-Maruyama) by evaluating the average error at different noise levels across the Books dataset. Formally, we compute each of the terms in Eq. \eqref{eq:matrix_score} separately:
\begin{align}
    \operatorname{error}_{X}(\tau) &= \frac{1}{|\mathcal{G}|} \sum_{v \in \mathcal{G}} \frac{\| \hat{\matr{X}}^{(j)}(v, \tau) - \matr{X}(v) \|_F}{N(v) \cdot F} \\
    \operatorname{error}_{A}(\tau) &= \frac{1}{|\mathcal{G}|} \sum_{v \in \mathcal{G}} \frac{\| \hat{\matr{A}}^{(j)}(v, \tau) - \matr{A}(v) \|_F}{(N(v))^2}.
\end{align}

\begin{figure*}[h]
     \centering
     \begin{subfigure}[b]{0.47\textwidth}
        \centering
        \includegraphics[width=\textwidth]{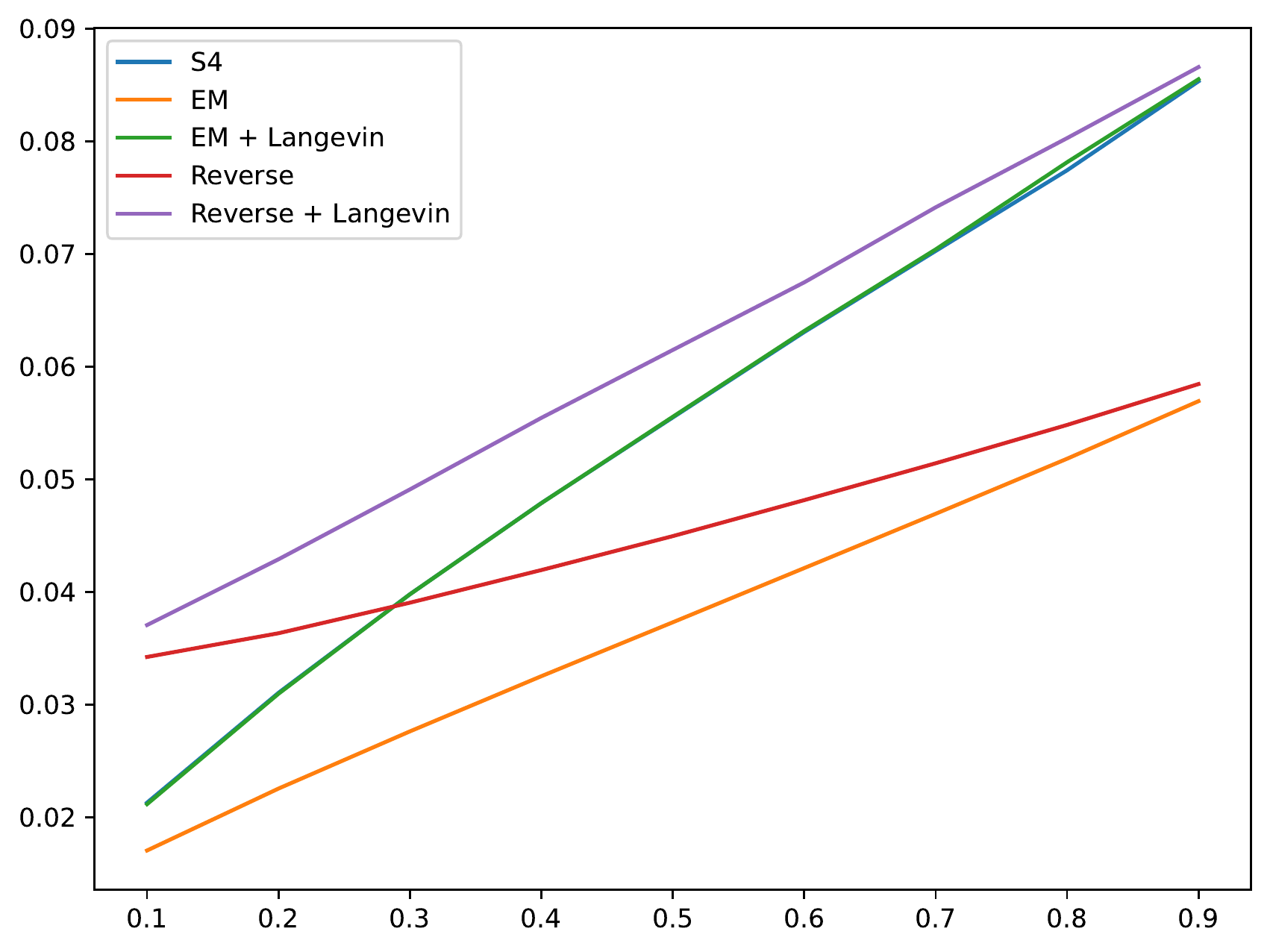}
        \caption{Feature matrix}
        \label{fig:x_err}
     \end{subfigure}
     \begin{subfigure}[b]{0.47\textwidth}
        \centering
        \includegraphics[width=\textwidth]{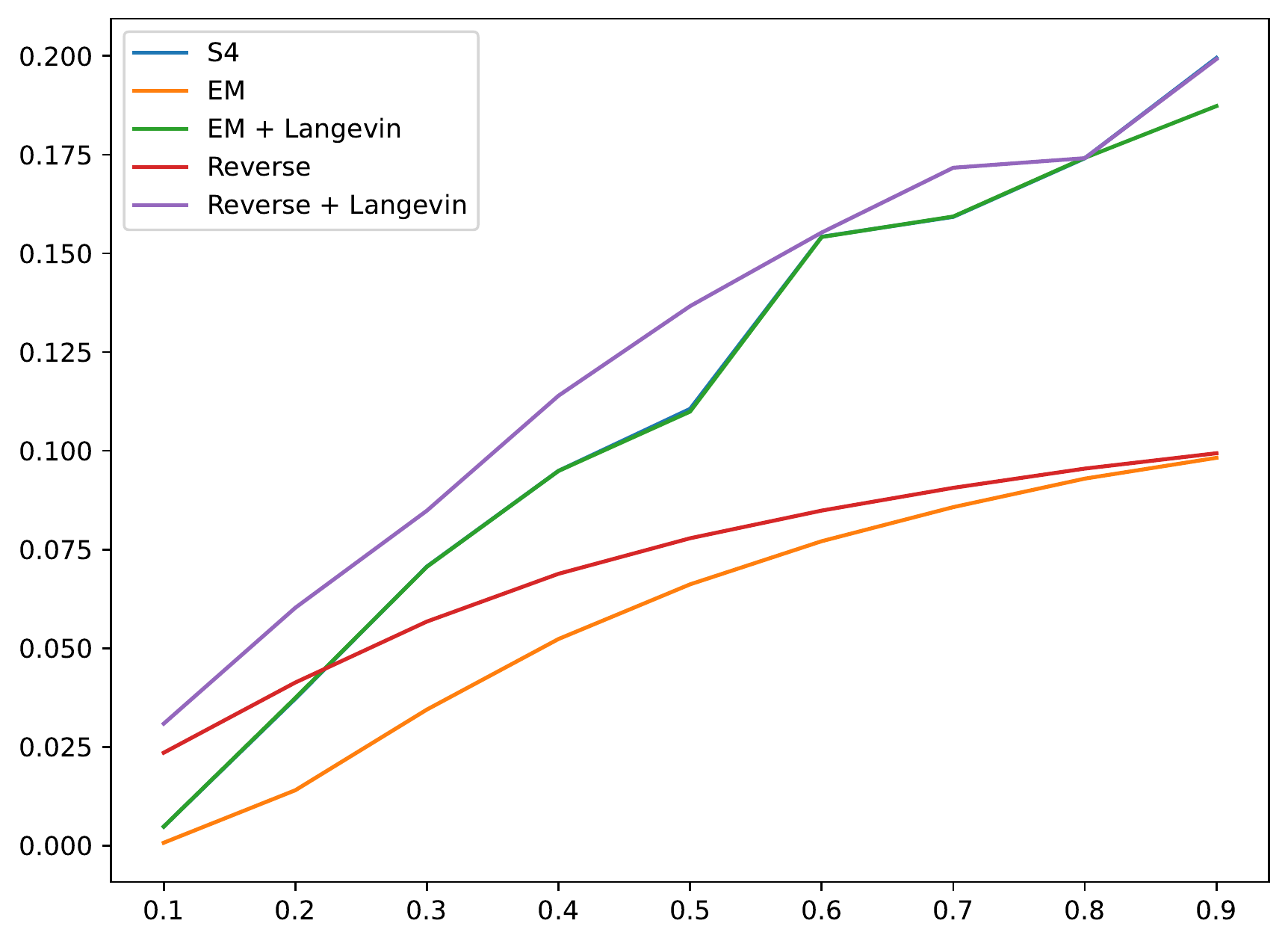}
        \caption{Adjacency matrix}
        \label{fig:adj_err}
     \end{subfigure}
        \caption{The reconstruction errors of each graph component ($y$-axis) vs time $\tau$ ($x$-axis)}
        \label{fig:reconstruction_errors}
\end{figure*}

Figure \ref{fig:reconstruction_errors} illustrates $\operatorname{error}_X$ and $\operatorname{error}_A$ for each solver at different reconstruction times $\tau$. The reported errors are averaged across 20 runs with different hyperparameters. As can be seen from the plots, a simple Euler-Maruyama (EM) scheme consistently outperforms all the other solvers.

Both \emph{EM} and \emph{Reverse} \cite{sde-score} correspond to different finite step discretization schemes of the underlying reverse-time SDE. The remaining methods (\emph{S4} \cite{sde-graph}, \emph{EM + Langevin}, \emph{Reverse + Langevin}) are Predictor-Corrector solvers that leverage the Langevin MCMC as a corrector to improve the quality of intermediate samples \cite{sde-score}.

\newpage
\section{Energy Histograms}
\label{appendix:energy}

\begin{figure*}[h!]
     \centering
     \includegraphics[width=0.95\textwidth]{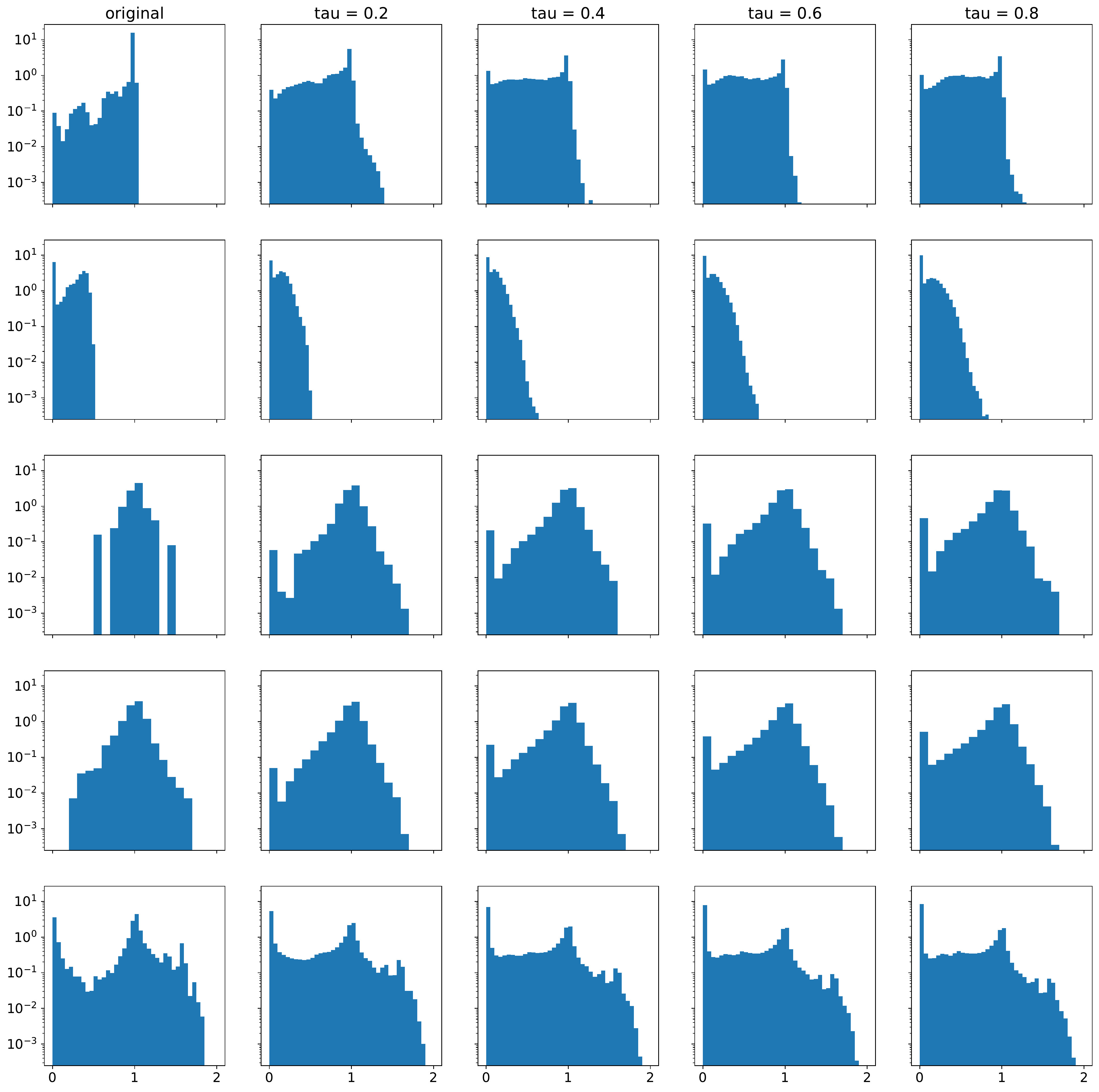}
     \caption{Histograms of the original and reconstructed normalized energies (log-densities). The datasets order from top to bottom: \emph{Weibo}, \emph{Reddit}, \emph{Disney}, \emph{Books}, \emph{Enron}.}
     \label{fig:energy}
\end{figure*}

\newpage

\begin{figure*}[h!]
     \centering
     \includegraphics[width=0.95\textwidth]{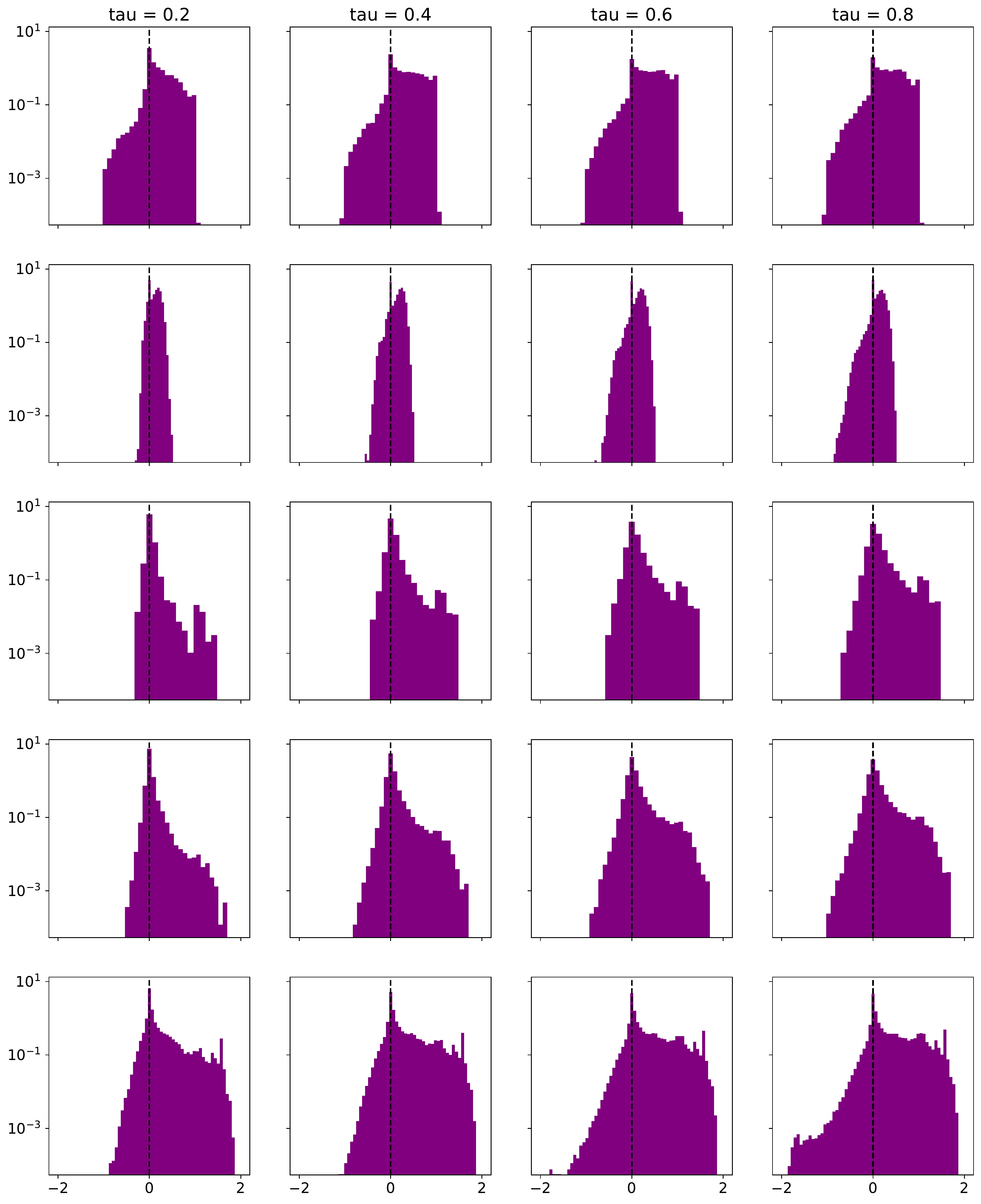}
     \caption{Histograms of the differences between the original and reconstructed normalized energies (log-densitites). Dashed black line is set to mark zero difference. The datasets order from top to bottom: \emph{Weibo}, \emph{Reddit}, \emph{Disney}, \emph{Books}, \emph{Enron}.}
     \label{fig:diff_energy}
\end{figure*}


\section{Ego-graphs}
\label{appendix:graphs}

In this appendix, we plot 8 randomly selected ego-graphs and their reconstructions from each dataset. We reconstruct ego-graphs using checkpoints with index 0 (there are overall 20 checkpoints for each network). Graphs are summarized by the number of nodes and edges, denoted as n and e, respectively. Node color indicates the relative error in features, with \textcolor{mycyan}{cyan} corresponding to perfect reconstruction, and \textcolor{mymagenta}{magenta} signifying that the error is comparable to the norm of features. The colors are interpolated using the \emph{cool} colormap from Matplotlib \cite{matplotlib}. The graph layout is computed using the Fruchterman-Reingold force-directed algorithm \cite{fruchterman1991graph} (\emph{spring\_layout} in NetworkX package \cite{networkx}).

\begin{figure*}[h]
     \centering
     \begin{subfigure}[b]{0.19\textwidth}
        \centering
        \includegraphics[width=\textwidth]{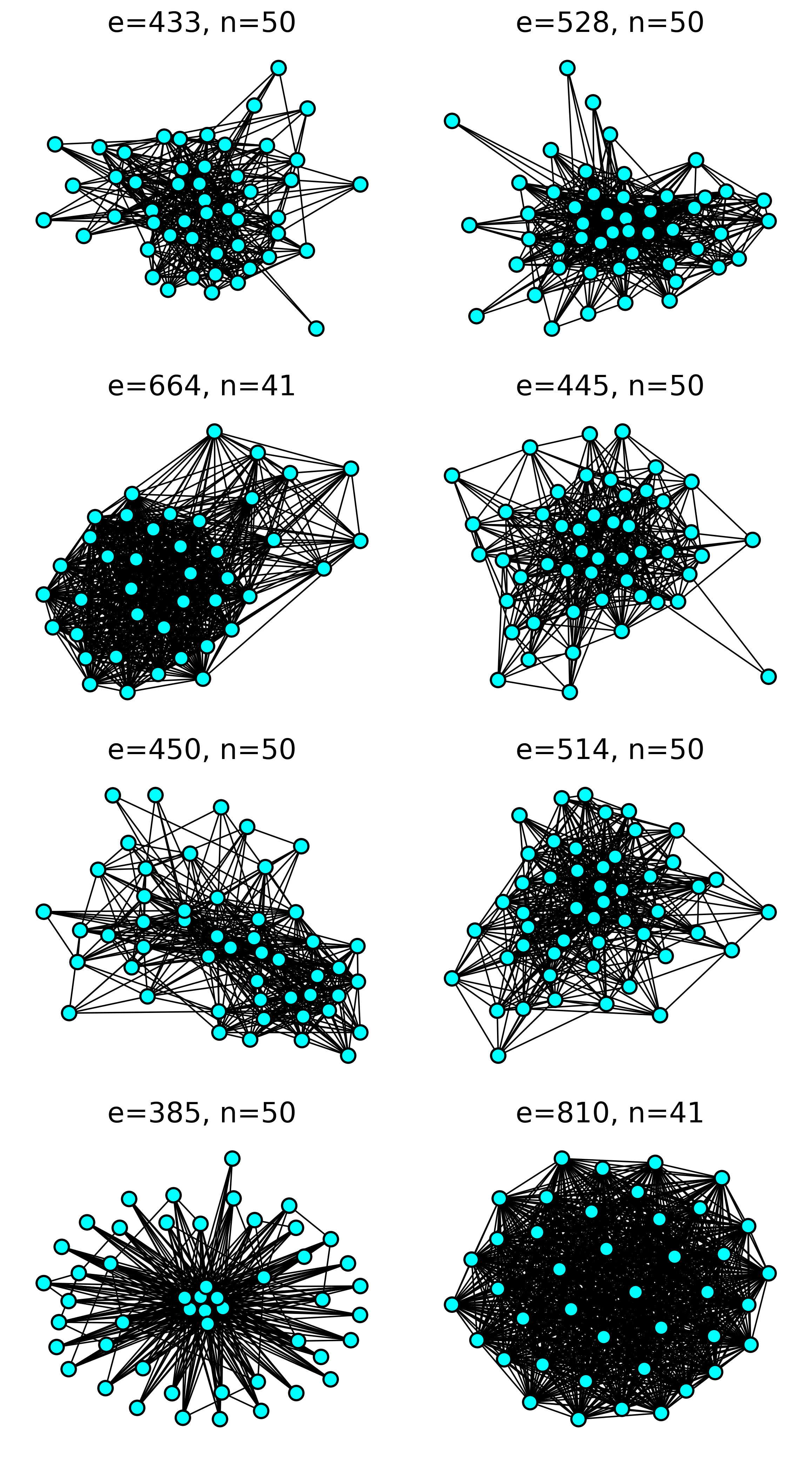}
        \caption{Original}
     \end{subfigure}
     \begin{subfigure}[b]{0.19\textwidth}
        \centering
        \includegraphics[width=\textwidth]{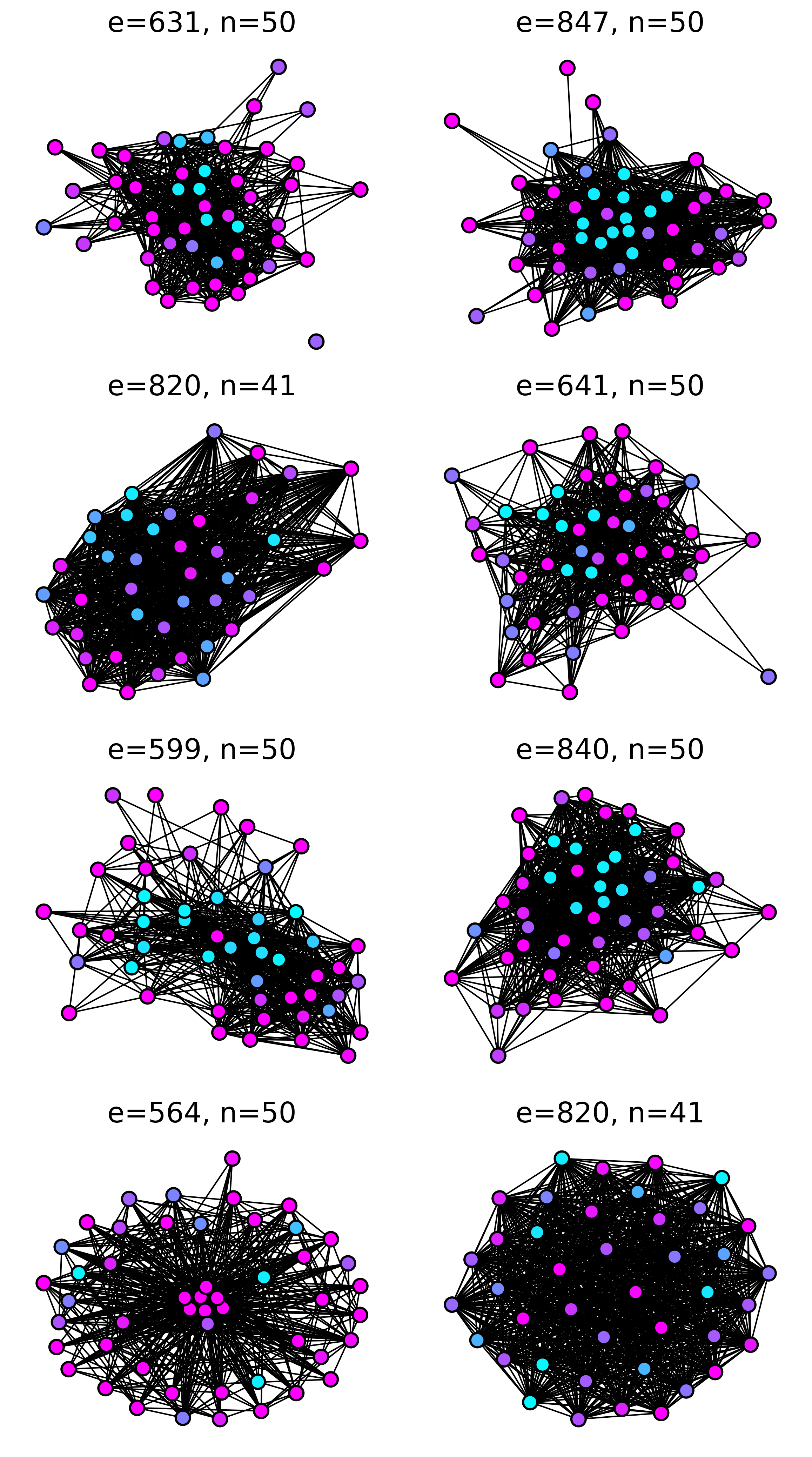}
        \caption{$\tau=0.2$}
     \end{subfigure}
     \begin{subfigure}[b]{0.19\textwidth}
        \centering
        \includegraphics[width=\textwidth]{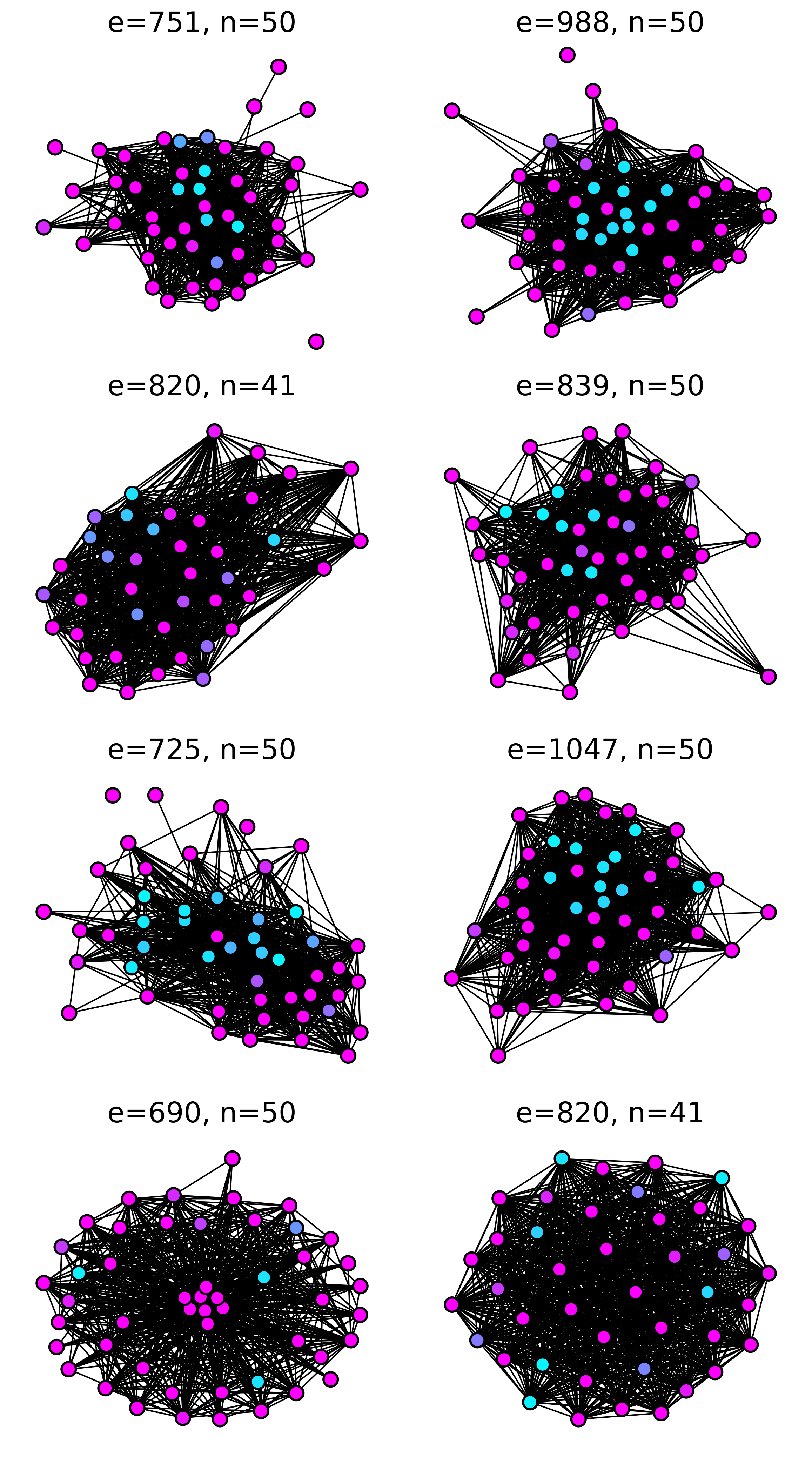}
        \caption{$\tau=0.4$}
     \end{subfigure}
     \begin{subfigure}[b]{0.19\textwidth}
        \centering
        \includegraphics[width=\textwidth]{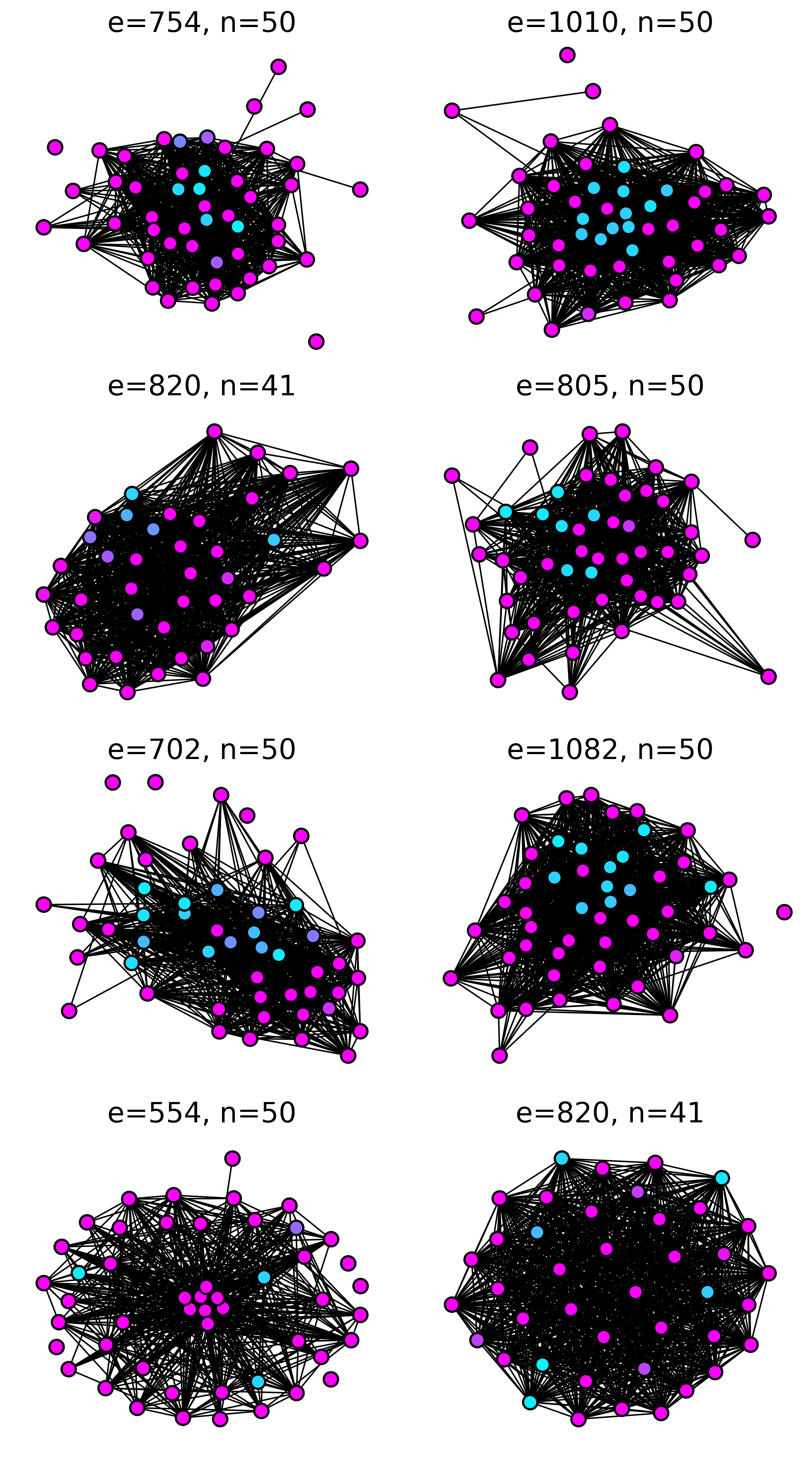}
        \caption{$\tau=0.6$}
     \end{subfigure}
     \begin{subfigure}[b]{0.19\textwidth}
        \centering
        \includegraphics[width=\textwidth]{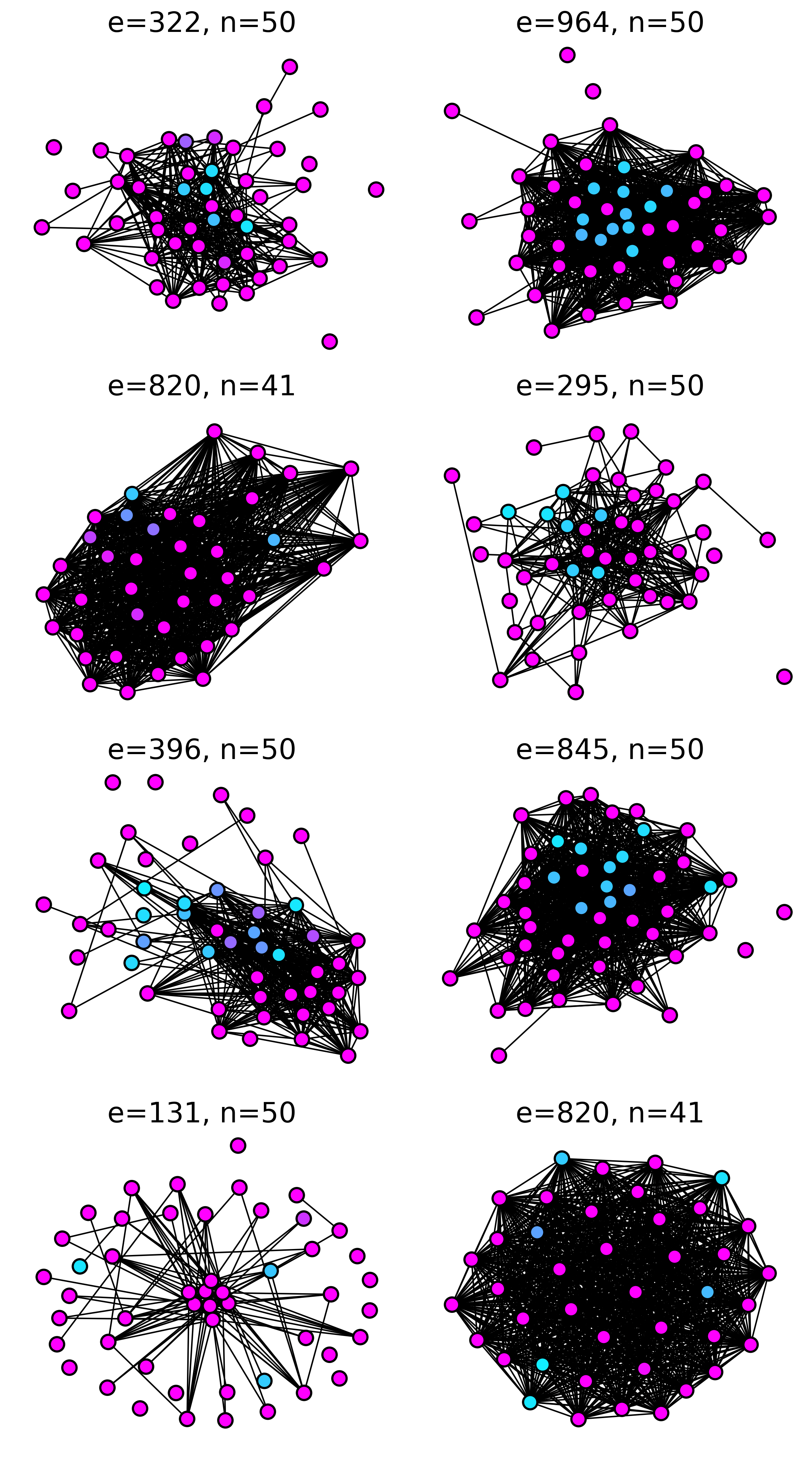}
        \caption{$\tau=0.8$}
     \end{subfigure}
        \caption{The original and reconstructed ego-graphs from Weibo dataset.}
        \label{fig:weibo_graphs}
\end{figure*}

\begin{figure*}[h]
     \centering
     \begin{subfigure}[b]{0.19\textwidth}
        \centering
        \includegraphics[width=\textwidth]{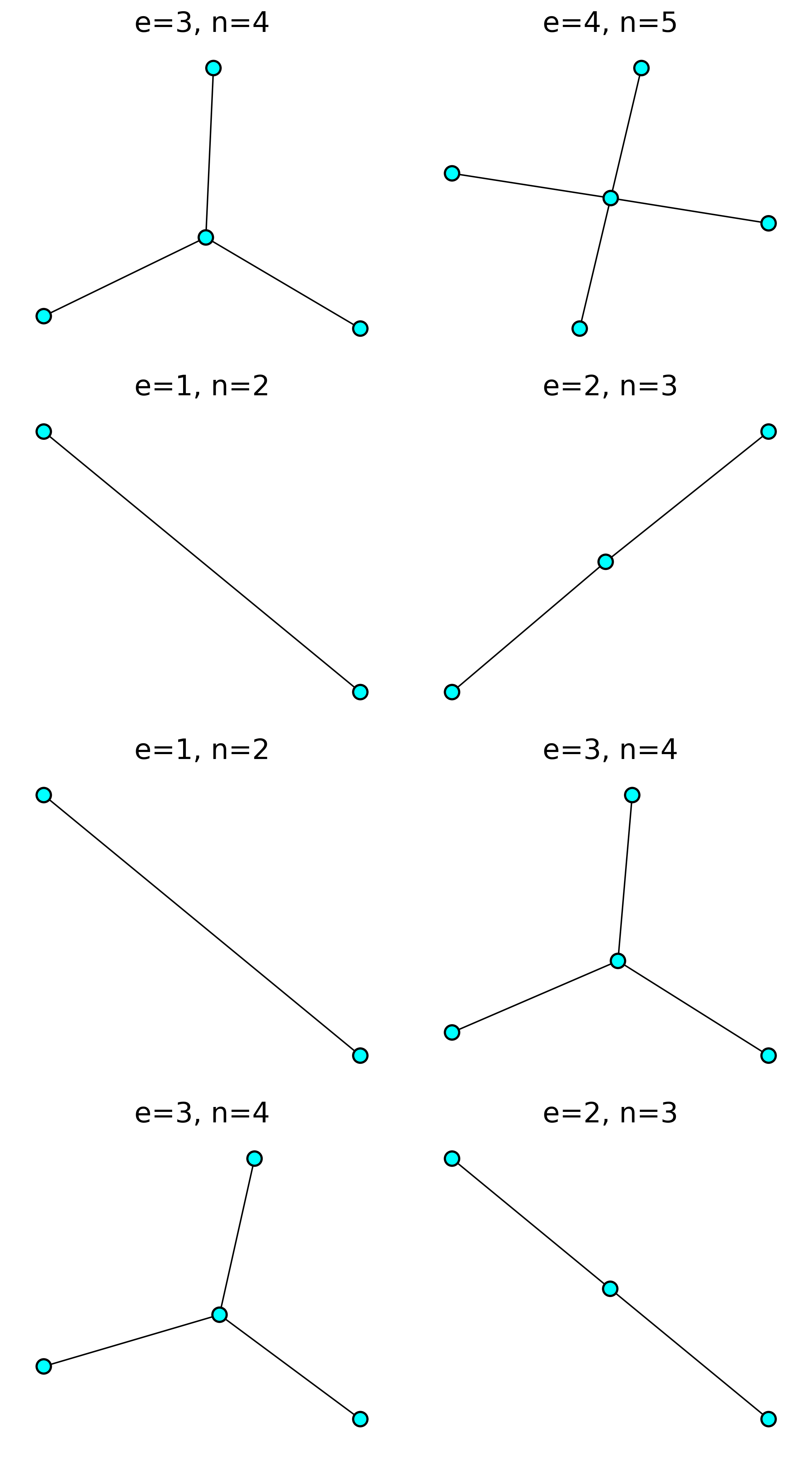}
        \caption{Original}
     \end{subfigure}
     \begin{subfigure}[b]{0.19\textwidth}
        \centering
        \includegraphics[width=\textwidth]{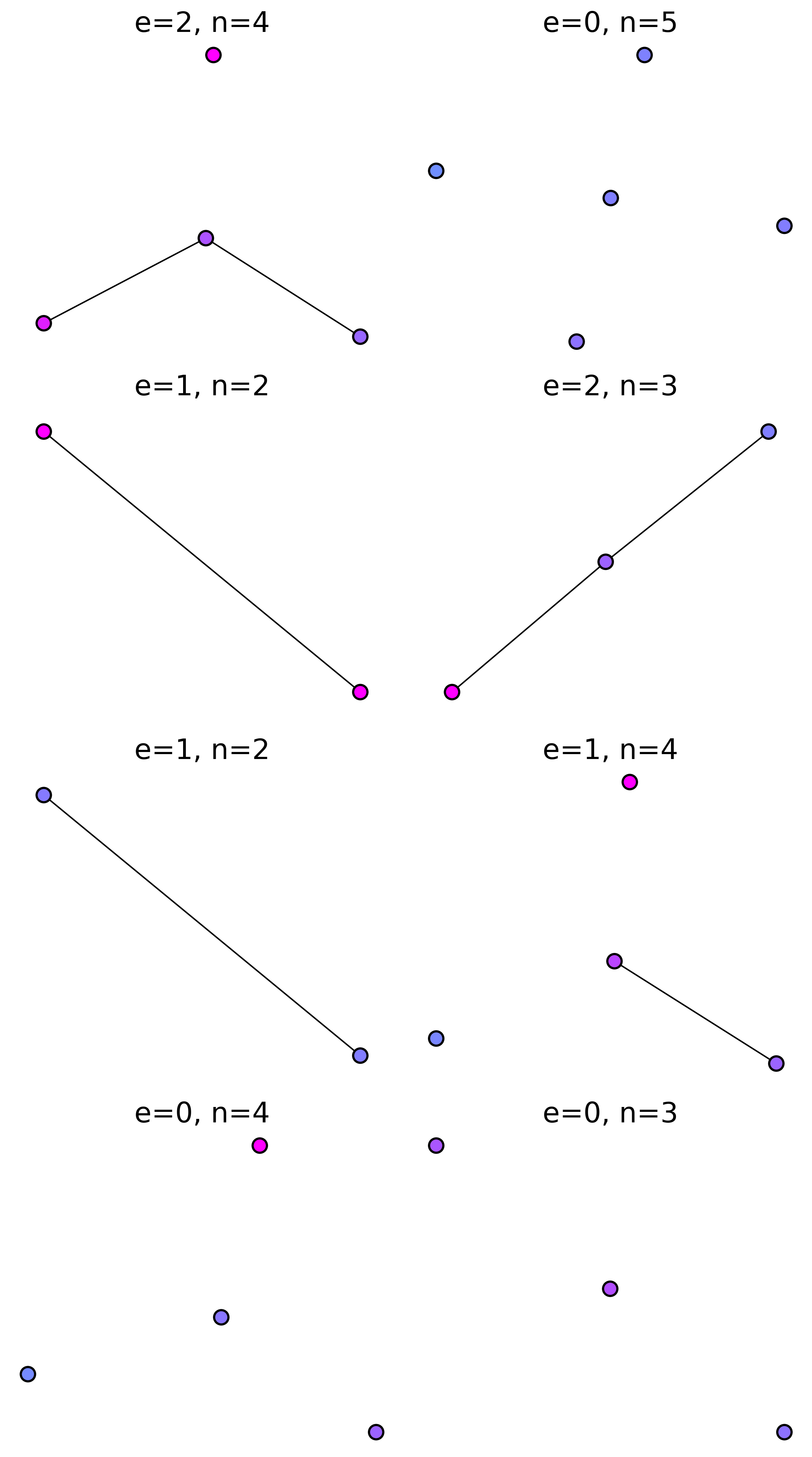}
        \caption{$\tau=0.2$}
     \end{subfigure}
     \begin{subfigure}[b]{0.19\textwidth}
        \centering
        \includegraphics[width=\textwidth]{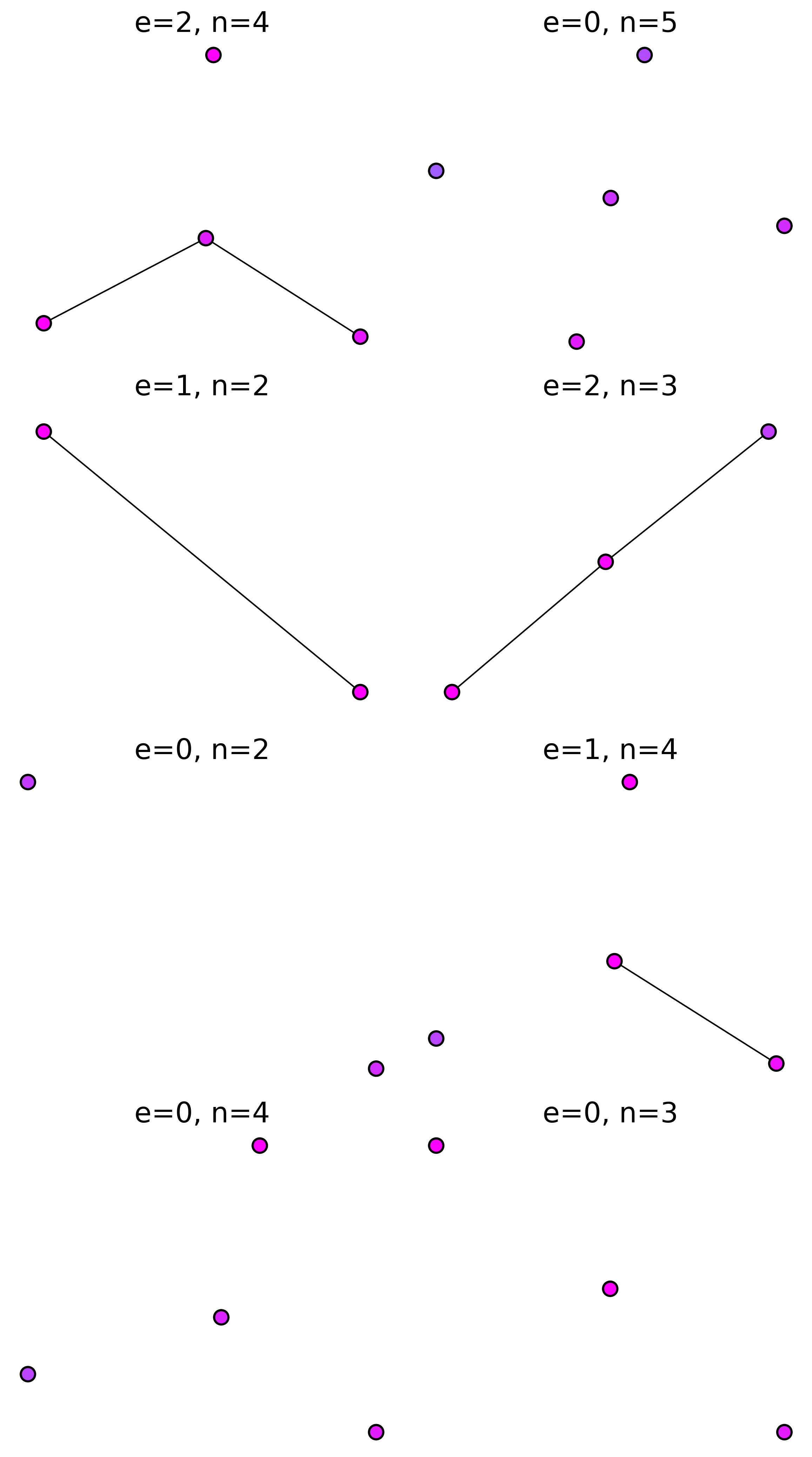}
        \caption{$\tau=0.4$}
     \end{subfigure}
     \begin{subfigure}[b]{0.19\textwidth}
        \centering
        \includegraphics[width=\textwidth]{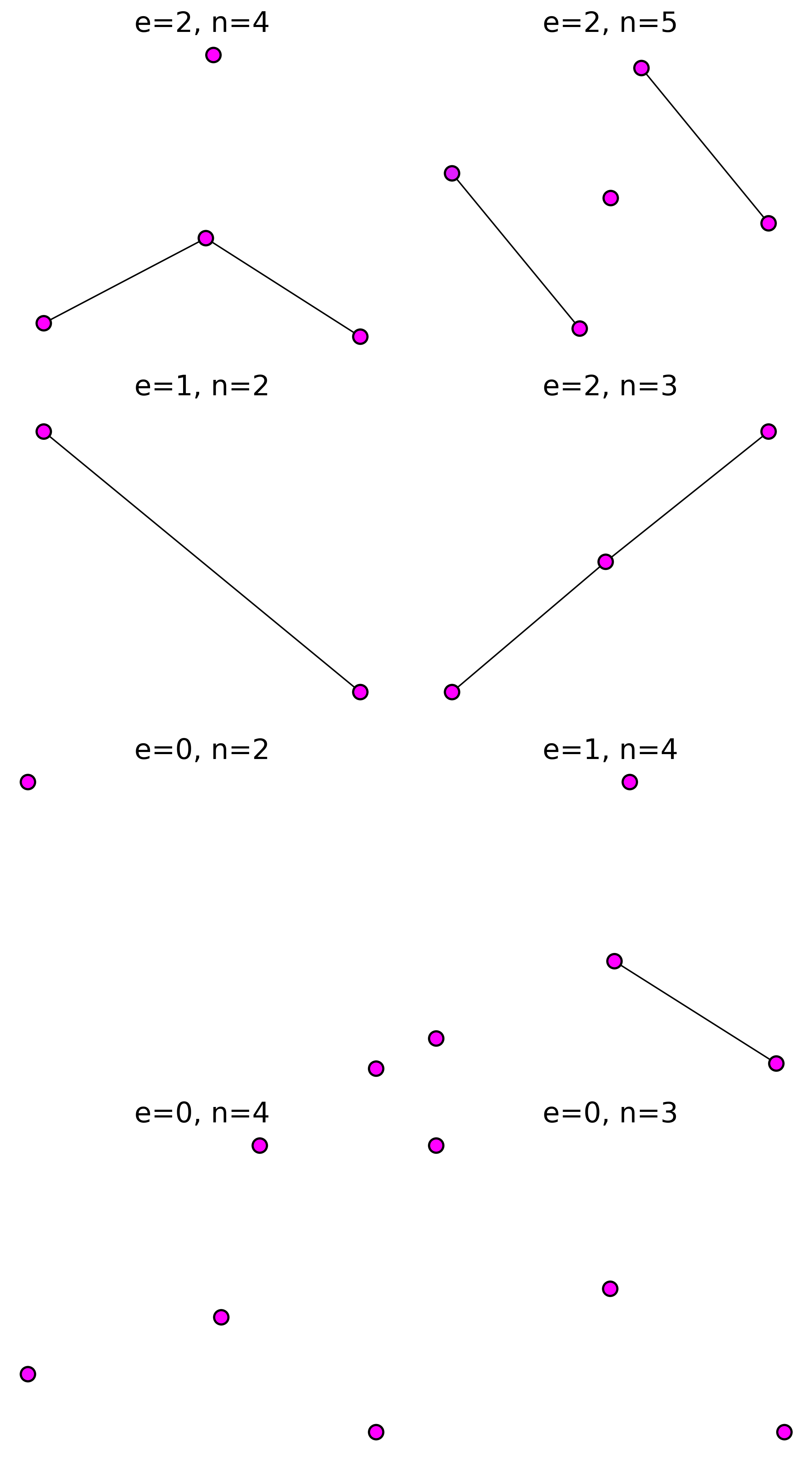}
        \caption{$\tau=0.6$}
     \end{subfigure}
     \begin{subfigure}[b]{0.19\textwidth}
        \centering
        \includegraphics[width=\textwidth]{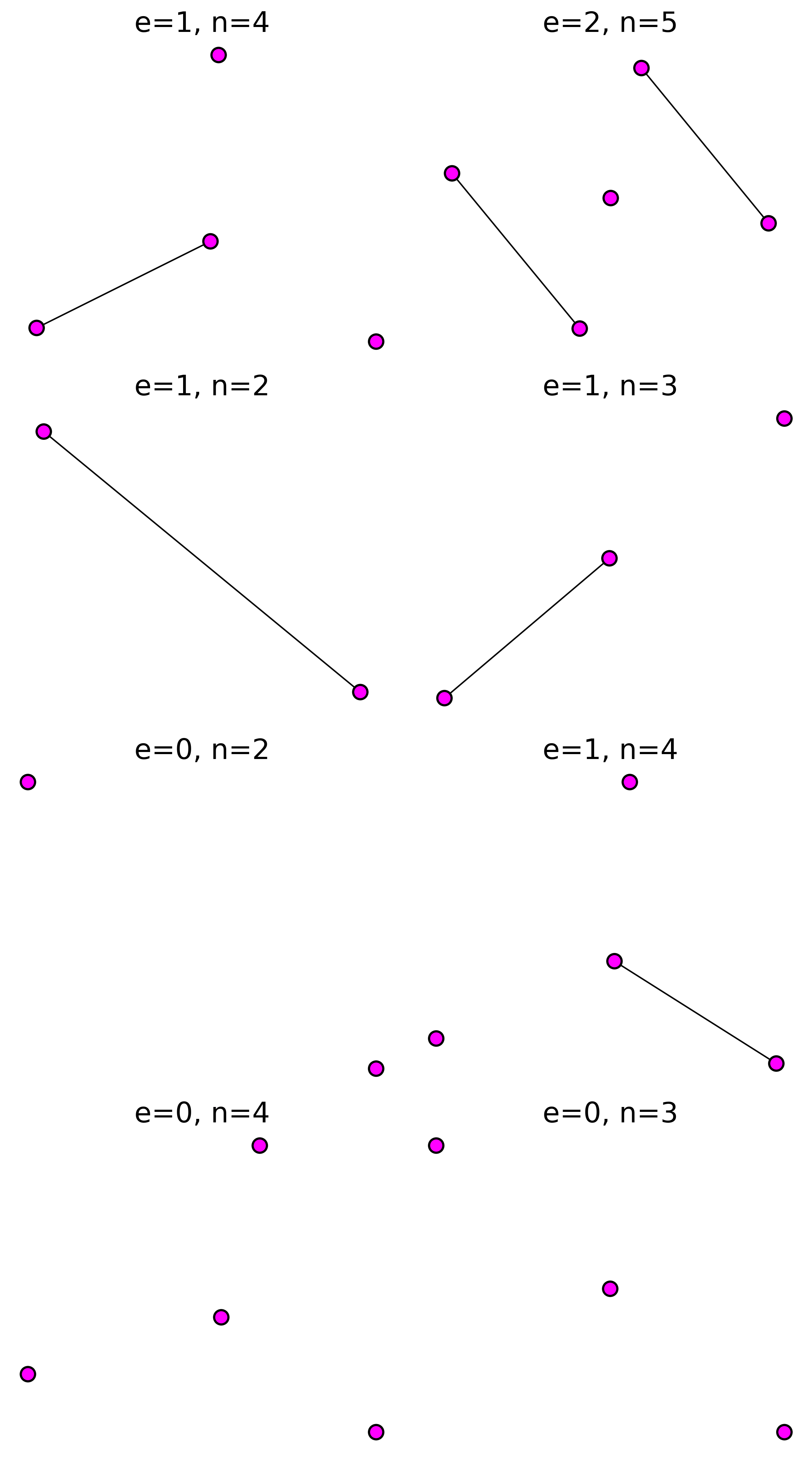}
        \caption{$\tau=0.8$}
     \end{subfigure}
        \caption{The original and reconstructed ego-graphs from Reddit dataset.}
        \label{fig:reddit_graphs}
\end{figure*}

\begin{figure*}[h]
     \centering
     \begin{subfigure}[b]{0.19\textwidth}
        \centering
        \includegraphics[width=\textwidth]{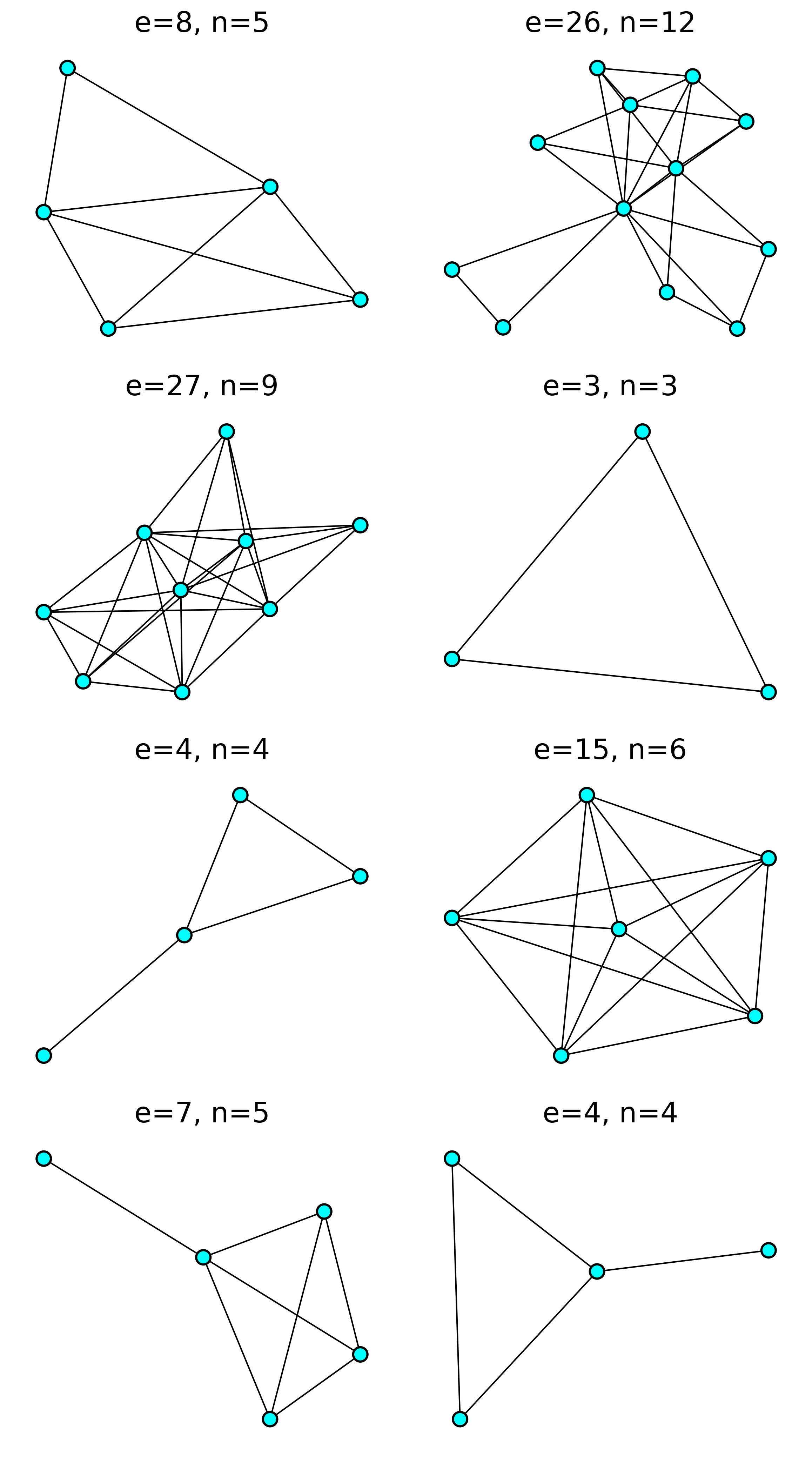}
        \caption{Original}
     \end{subfigure}
     \begin{subfigure}[b]{0.19\textwidth}
        \centering
        \includegraphics[width=\textwidth]{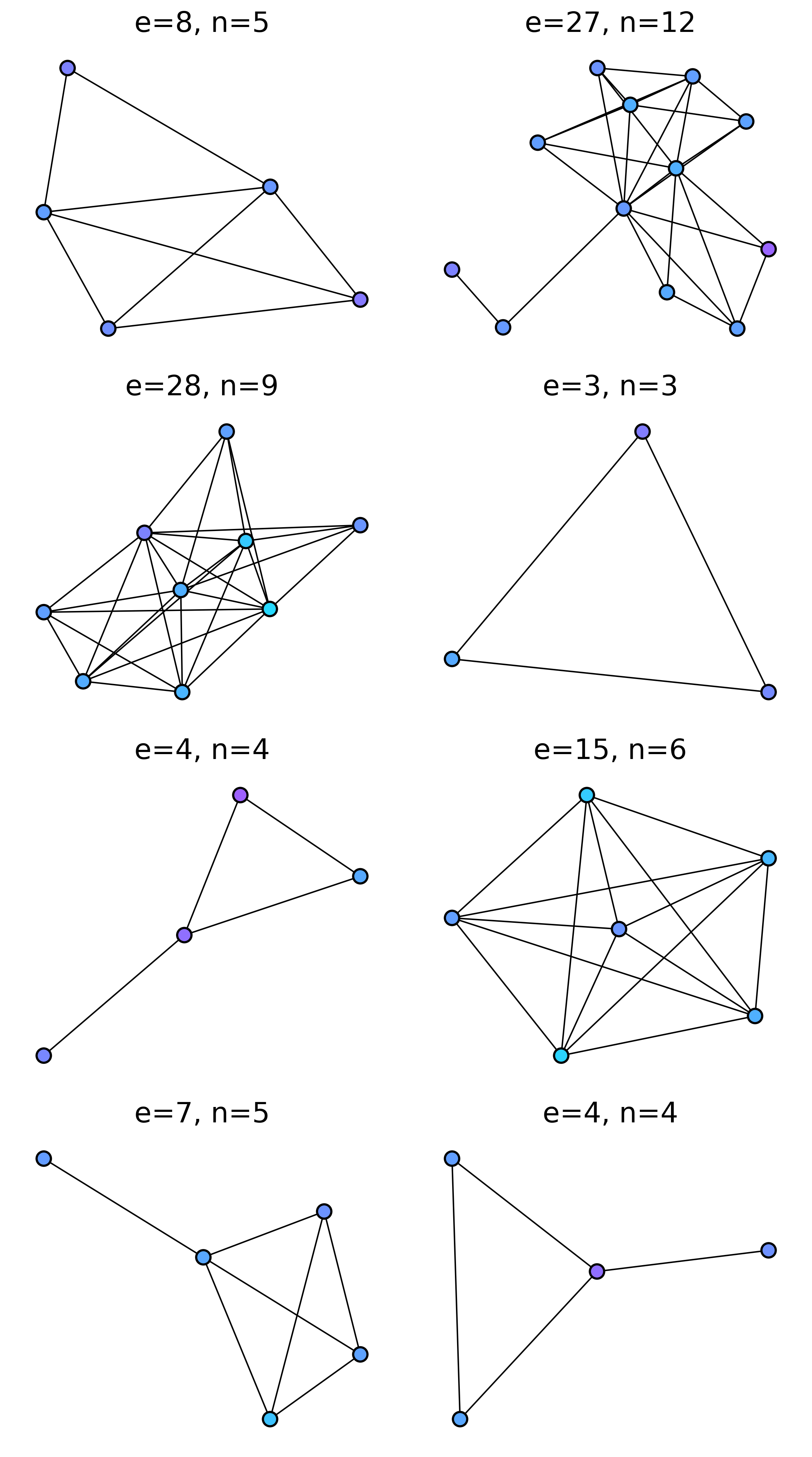}
        \caption{$\tau=0.2$}
     \end{subfigure}
     \begin{subfigure}[b]{0.19\textwidth}
        \centering
        \includegraphics[width=\textwidth]{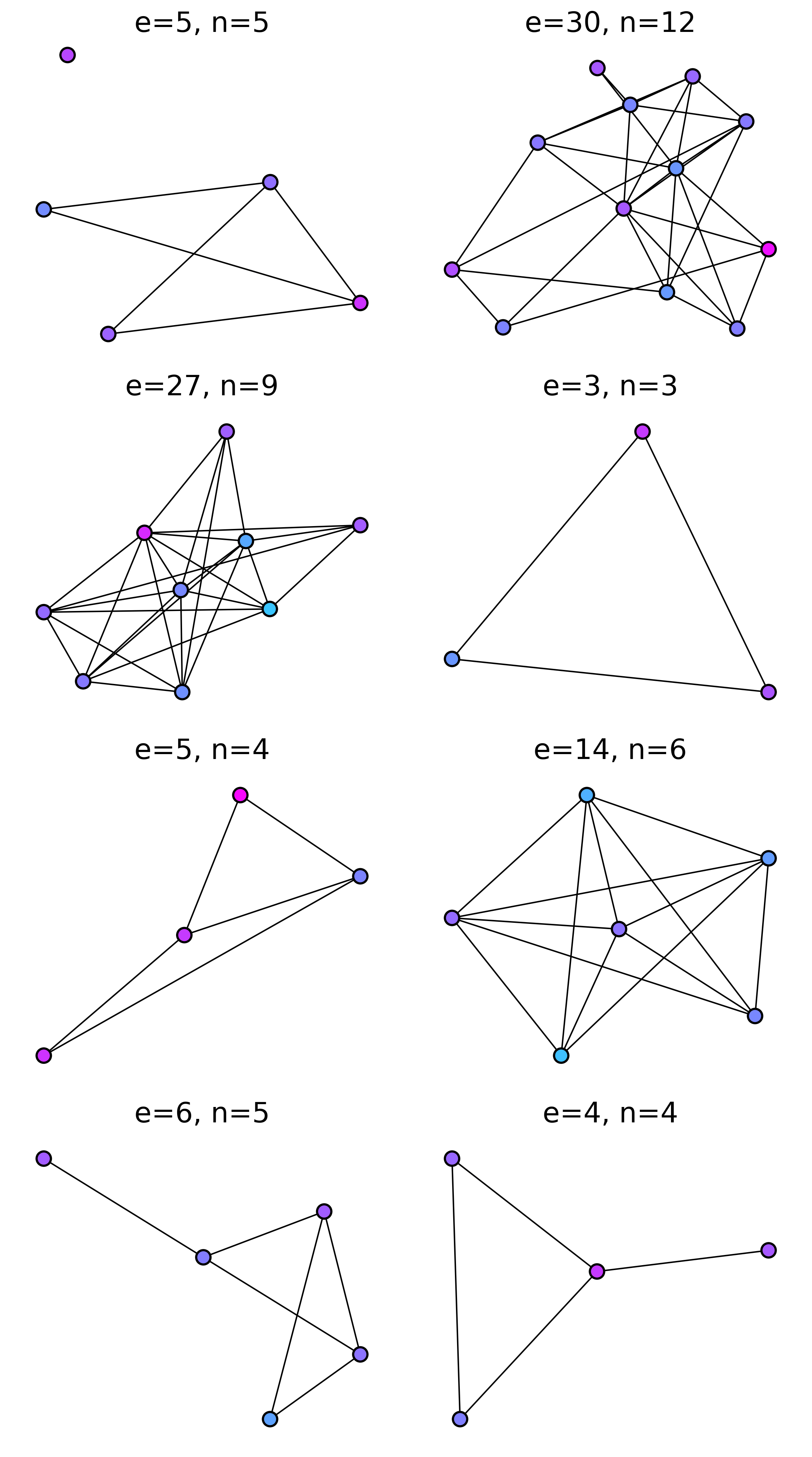}
        \caption{$\tau=0.4$}
     \end{subfigure}
     \begin{subfigure}[b]{0.19\textwidth}
        \centering
        \includegraphics[width=\textwidth]{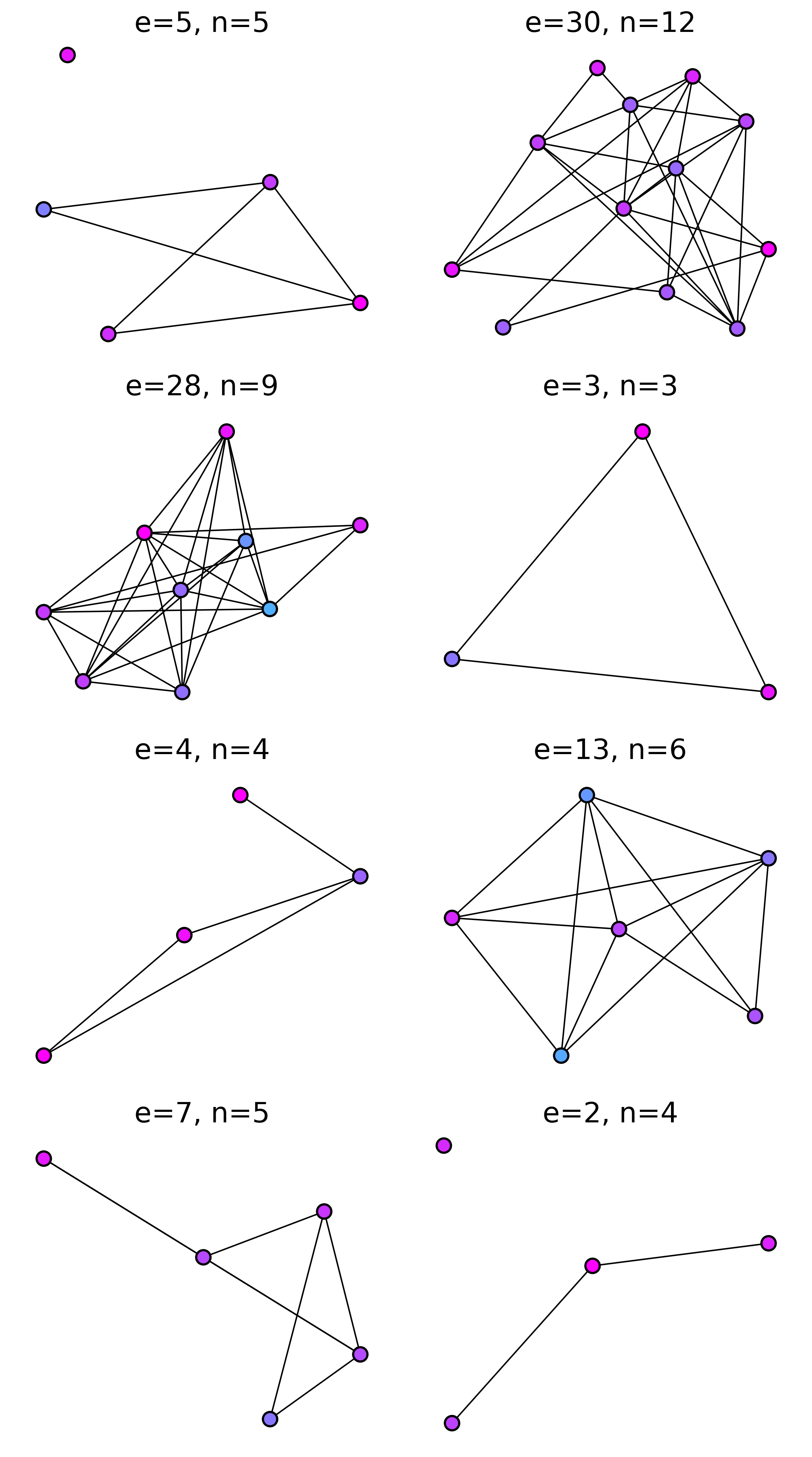}
        \caption{$\tau=0.6$}
     \end{subfigure}
     \begin{subfigure}[b]{0.19\textwidth}
        \centering
        \includegraphics[width=\textwidth]{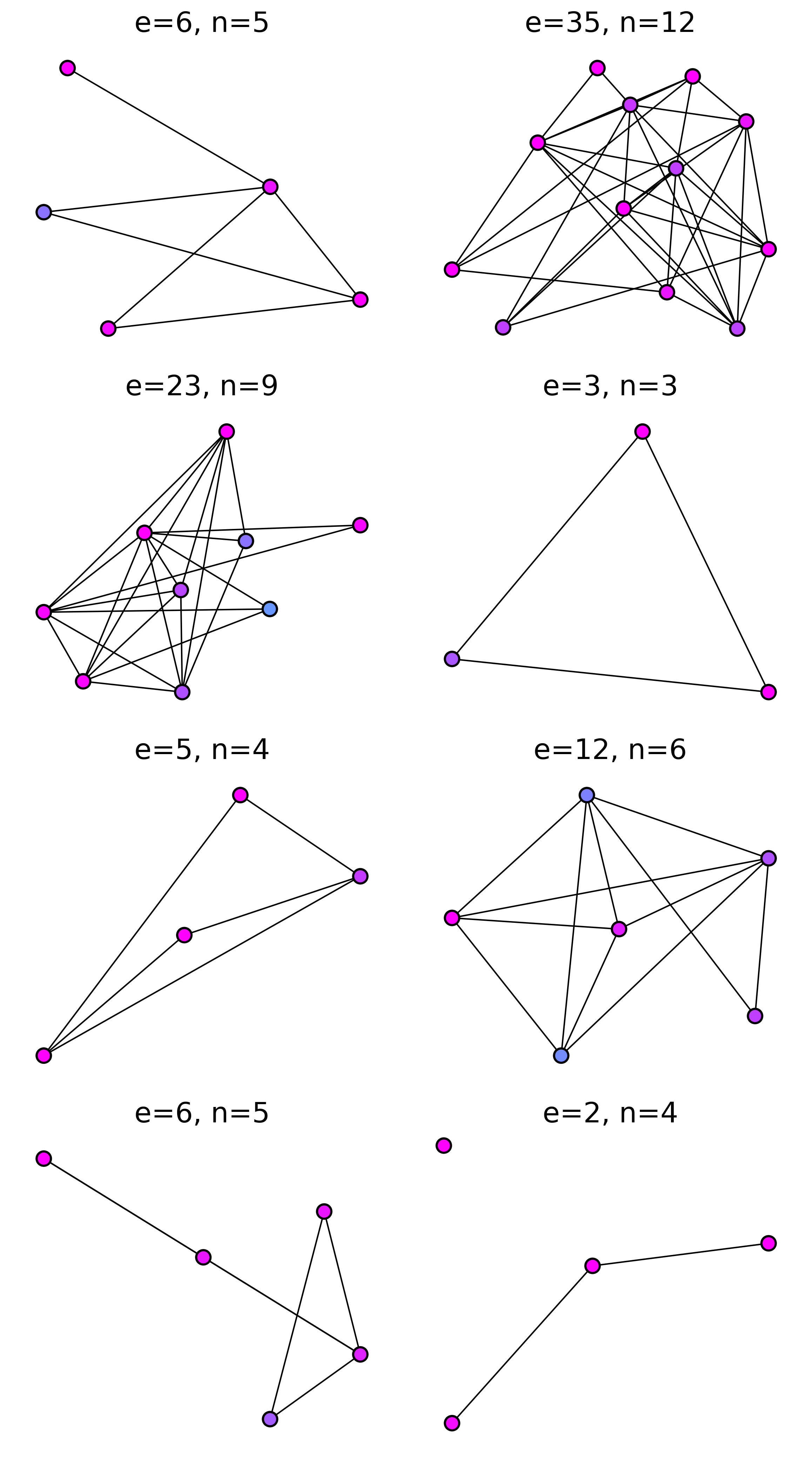}
        \caption{$\tau=0.8$}
     \end{subfigure}
        \caption{The original and reconstructed ego-graphs from Disney dataset.}
        \label{fig:disney_graphs}
\end{figure*}

\begin{figure*}[h]
     \centering
     \begin{subfigure}[b]{0.19\textwidth}
        \centering
        \includegraphics[width=\textwidth]{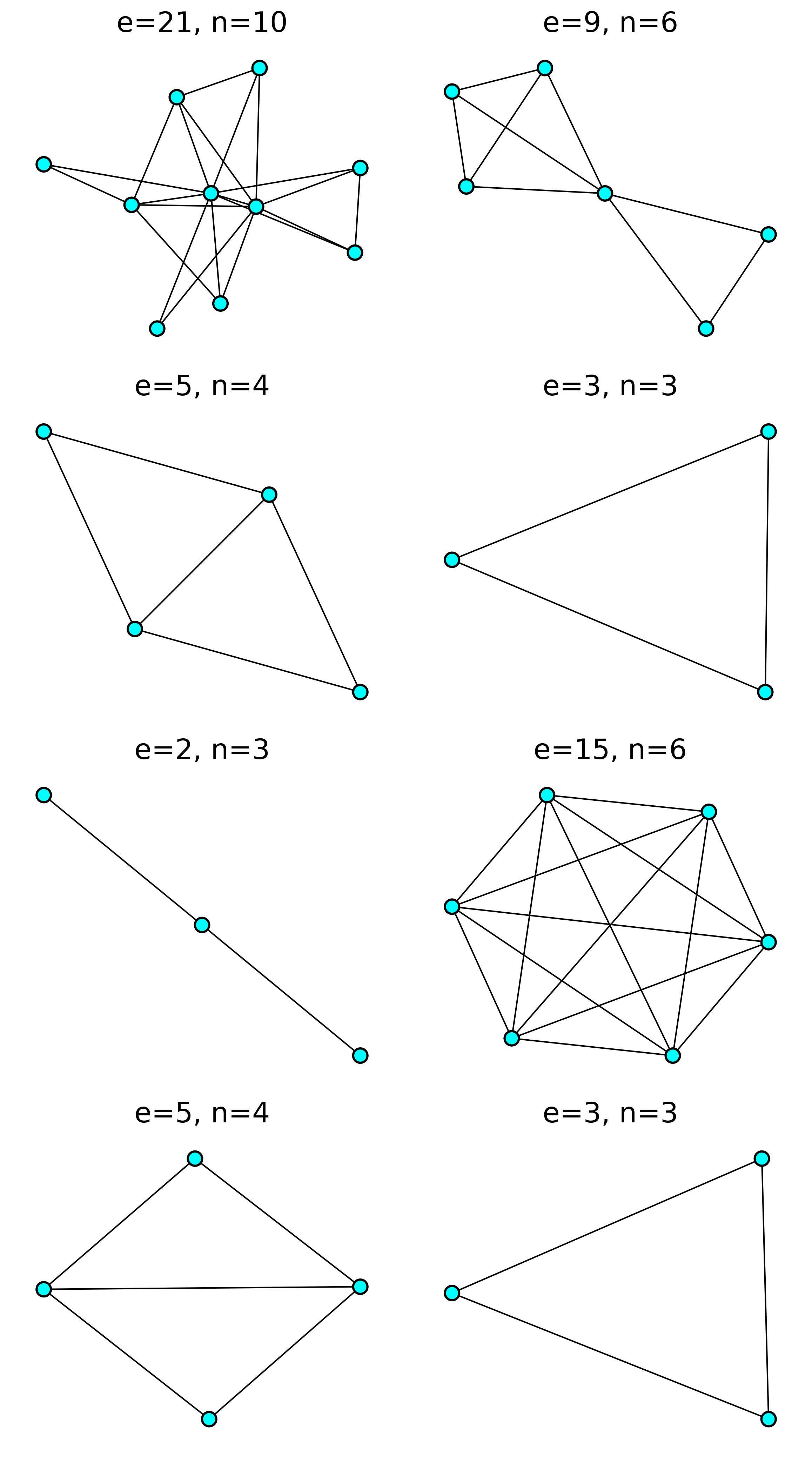}
        \caption{Original}
     \end{subfigure}
     \begin{subfigure}[b]{0.19\textwidth}
        \centering
        \includegraphics[width=\textwidth]{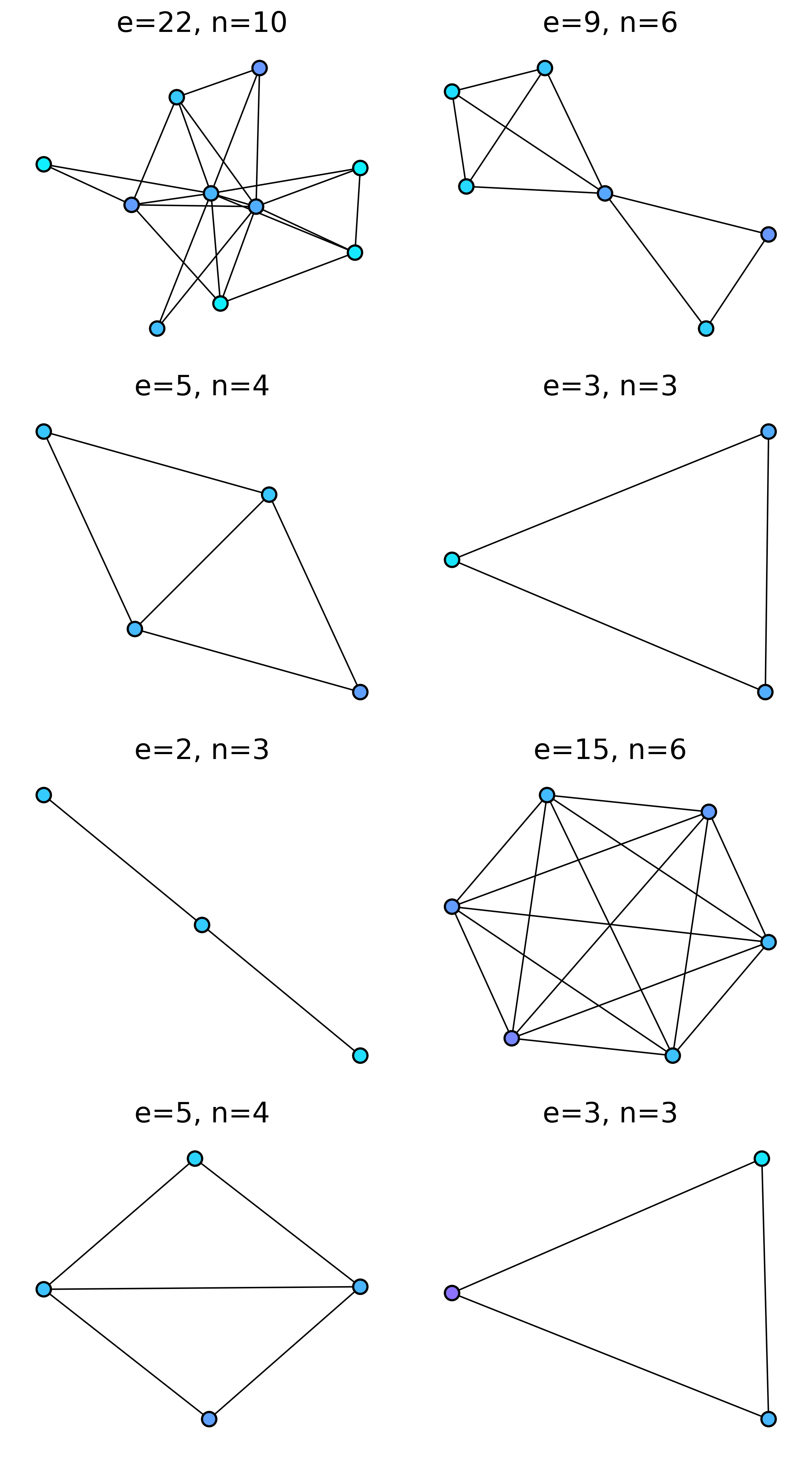}
        \caption{$\tau=0.2$}
     \end{subfigure}
     \begin{subfigure}[b]{0.19\textwidth}
        \centering
        \includegraphics[width=\textwidth]{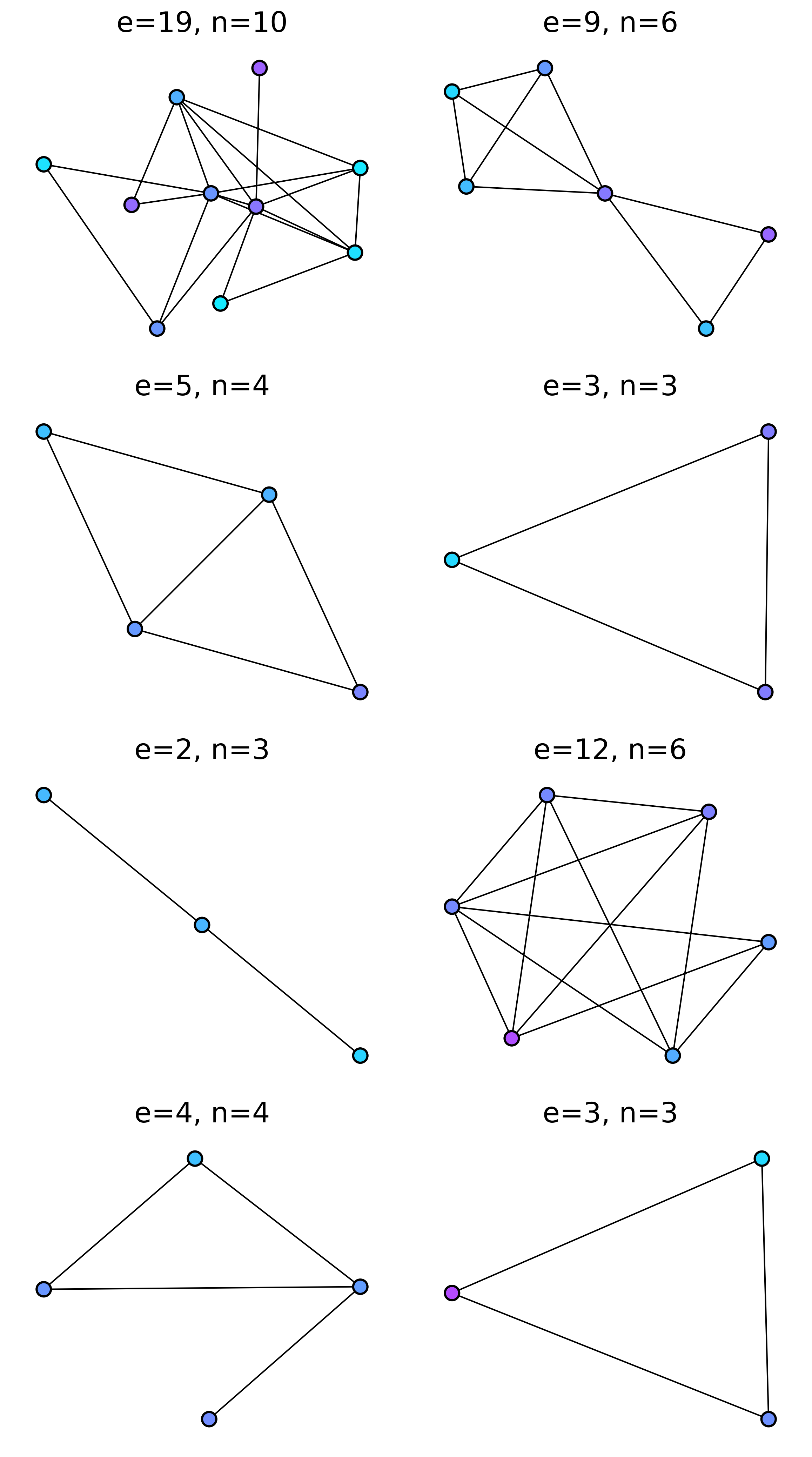}
        \caption{$\tau=0.4$}
     \end{subfigure}
     \begin{subfigure}[b]{0.19\textwidth}
        \centering
        \includegraphics[width=\textwidth]{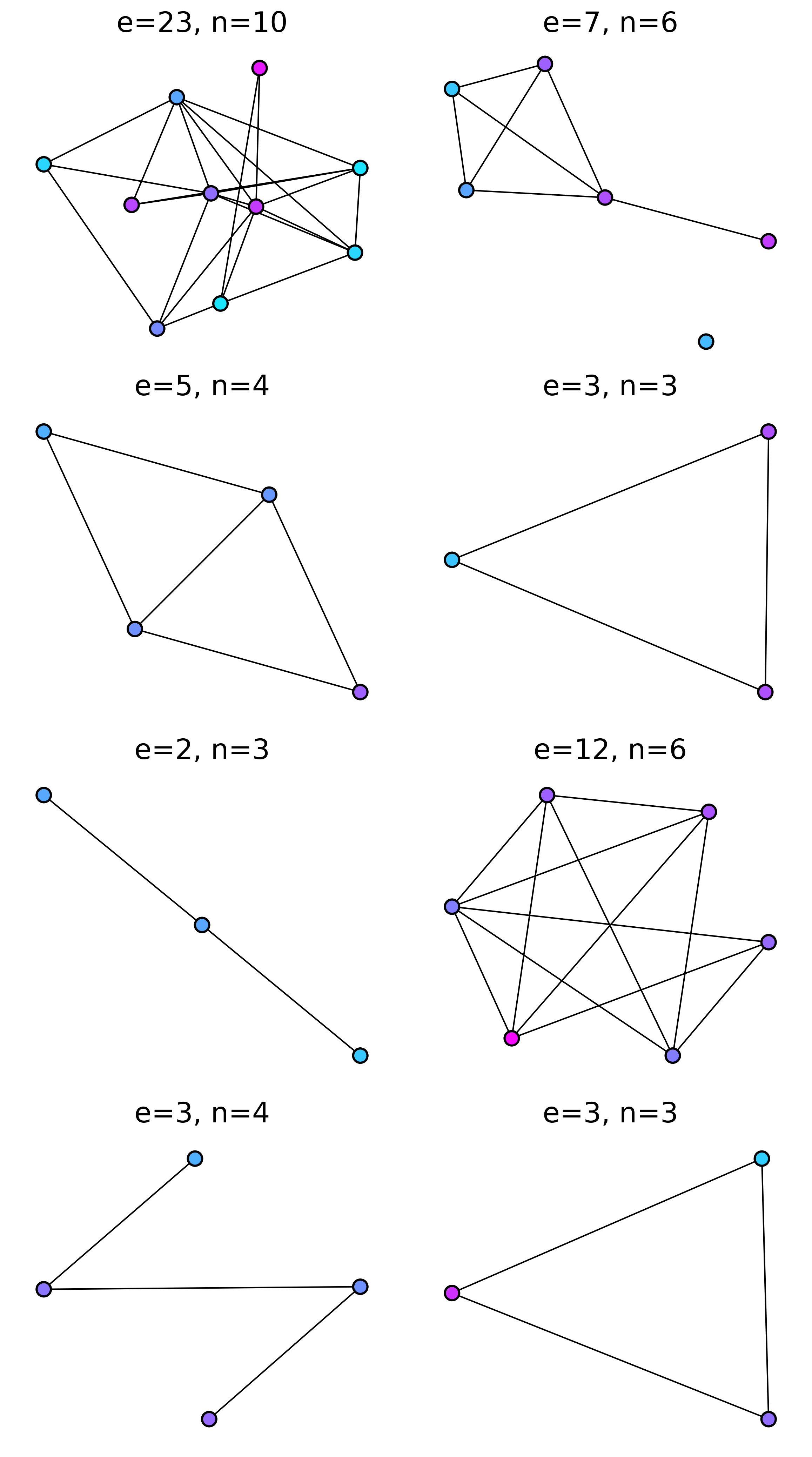}
        \caption{$\tau=0.6$}
     \end{subfigure}
     \begin{subfigure}[b]{0.19\textwidth}
        \centering
        \includegraphics[width=\textwidth]{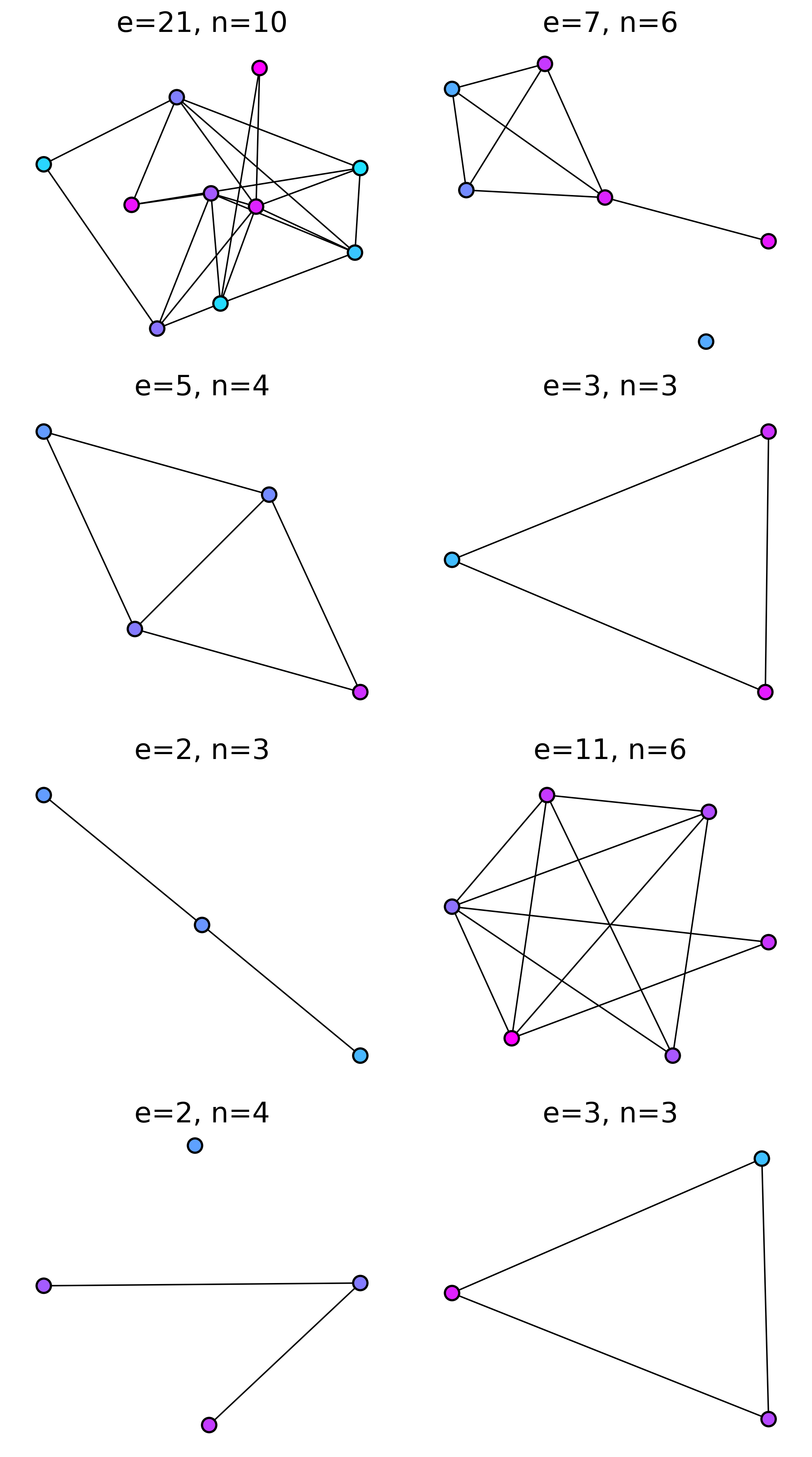}
        \caption{$\tau=0.8$}
     \end{subfigure}
        \caption{The original and reconstructed ego-graphs from Books dataset.}
        \label{fig:books_graphs}
\end{figure*}

\begin{figure*}[h]
     \centering
     \begin{subfigure}[b]{0.19\textwidth}
        \centering
        \includegraphics[width=\textwidth]{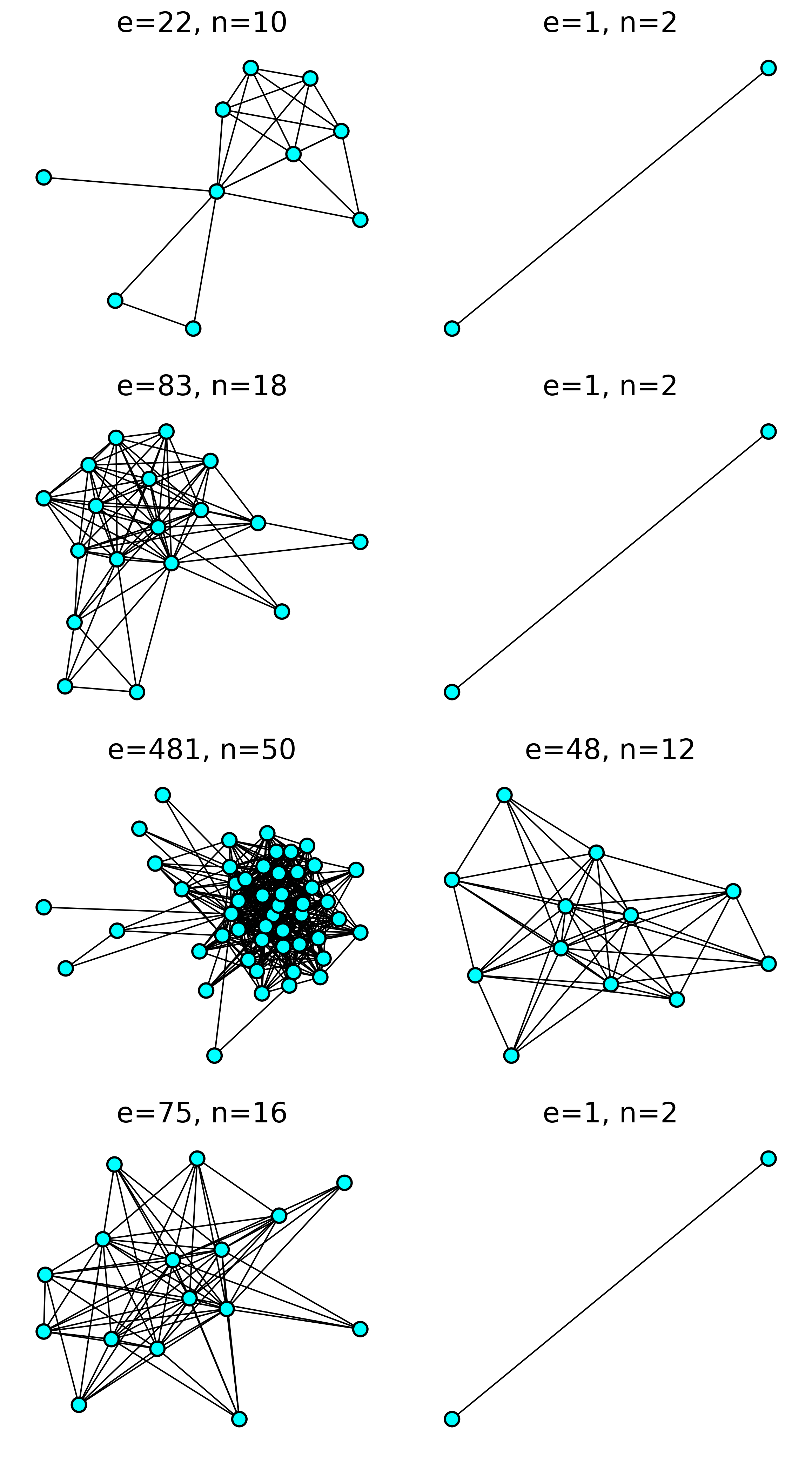}
        \caption{Original}
     \end{subfigure}
     \begin{subfigure}[b]{0.19\textwidth}
        \centering
        \includegraphics[width=\textwidth]{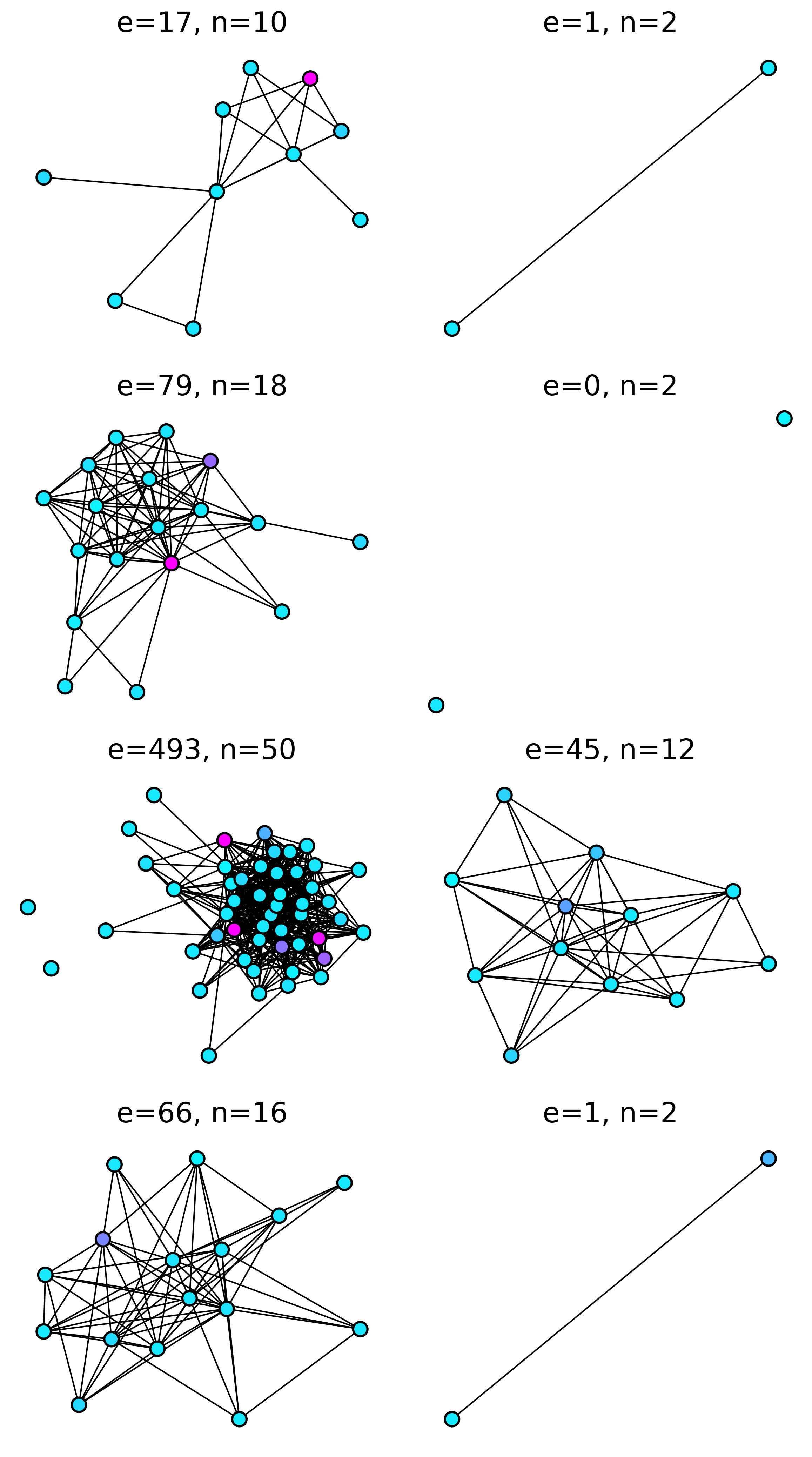}
        \caption{$\tau=0.2$}
     \end{subfigure}
     \begin{subfigure}[b]{0.19\textwidth}
        \centering
        \includegraphics[width=\textwidth]{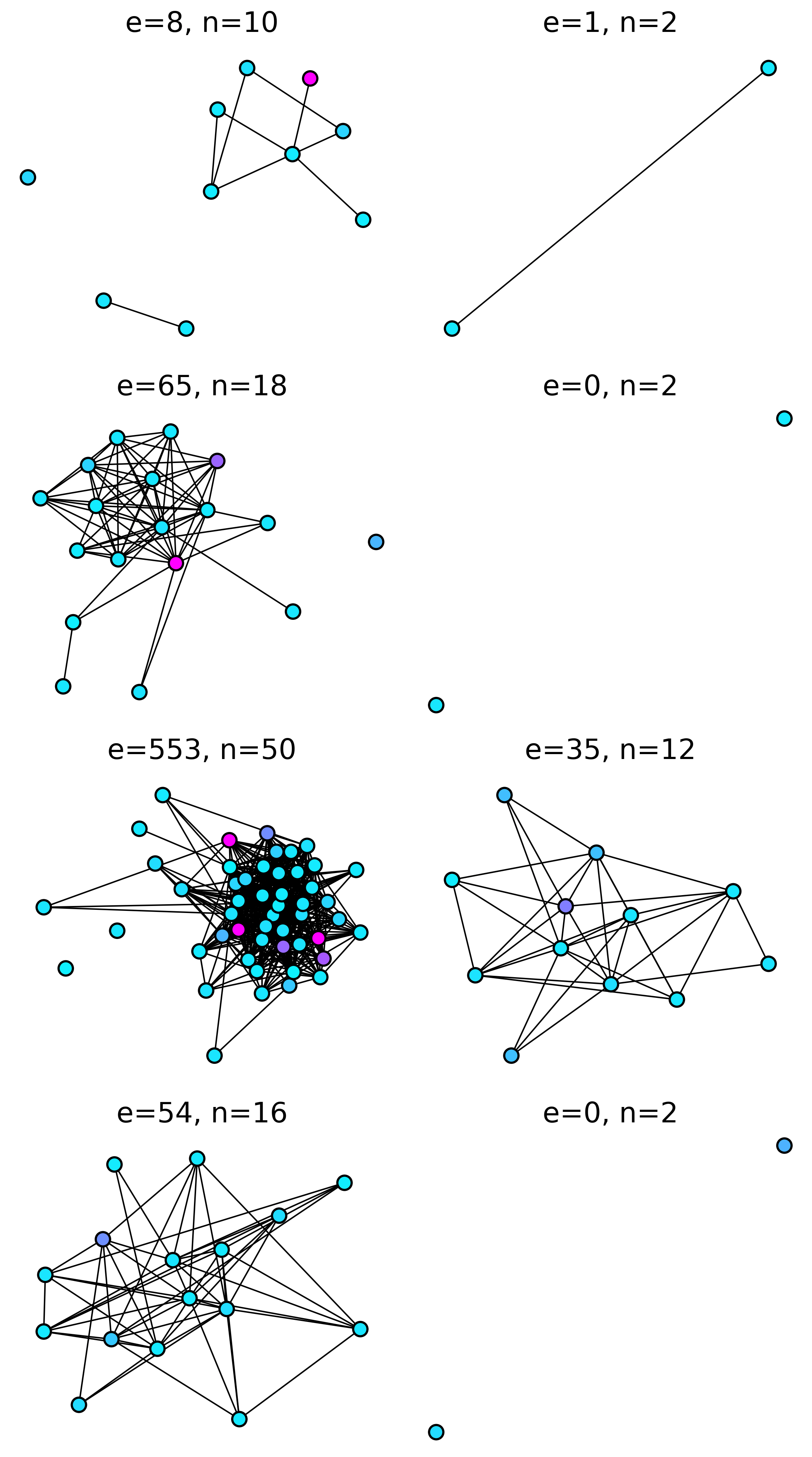}
        \caption{$\tau=0.4$}
     \end{subfigure}
     \begin{subfigure}[b]{0.19\textwidth}
        \centering
        \includegraphics[width=\textwidth]{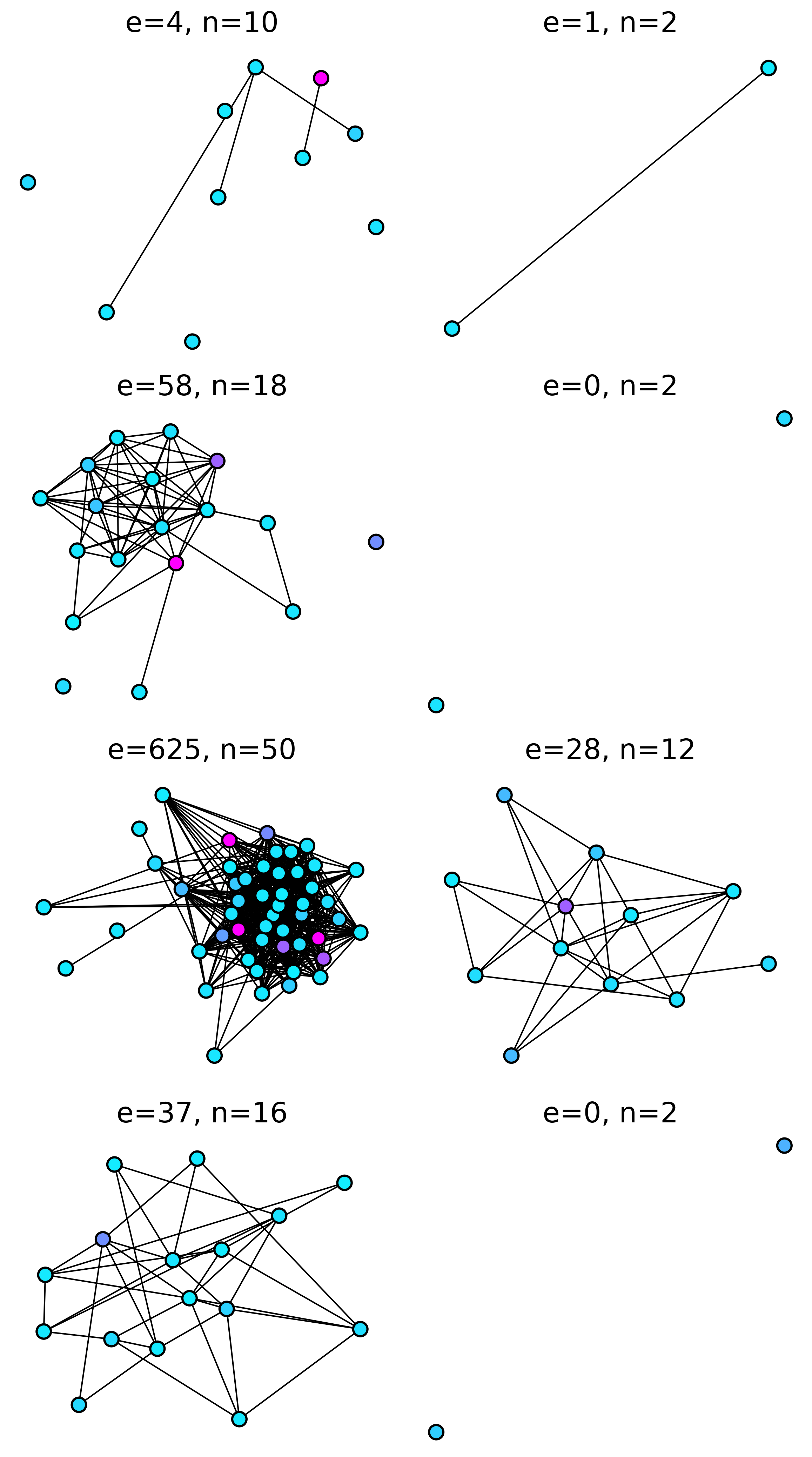}
        \caption{$\tau=0.6$}
     \end{subfigure}
     \begin{subfigure}[b]{0.19\textwidth}
        \centering
        \includegraphics[width=\textwidth]{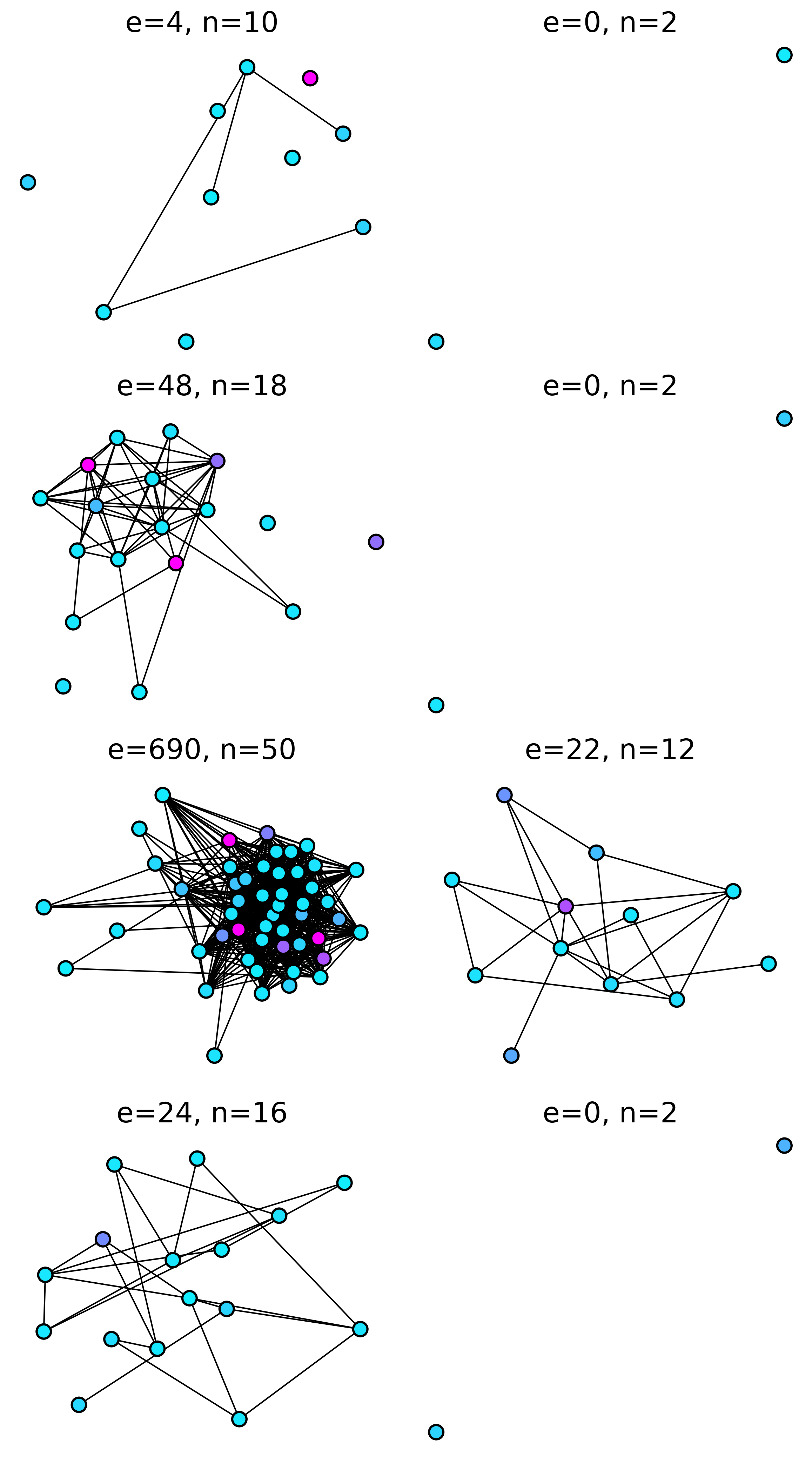}
        \caption{$\tau=0.8$}
     \end{subfigure}
        \caption{The original and reconstructed ego-graphs from Enron dataset.}
        \label{fig:enron_graphs}
\end{figure*}


\end{document}